\documentclass{article}

\usepackage{microtype}
\usepackage{graphicx}
\usepackage{subcaption}
\usepackage{booktabs}
\usepackage{tablefootnote}

\usepackage{colortbl}
\usepackage[normalem]{ulem}

\usepackage[hyphens]{url}
\usepackage{hyperref}
\usepackage[frozencache,cachedir=.]{minted} 
\usepackage{bbm}
\usepackage{xcolor}
\usepackage{amsmath}
\usepackage{multirow}
\usepackage{graphicx}
\usepackage{caption}
\usepackage{booktabs}
\usepackage{xcolor}
\usepackage{amsmath,amscd,amssymb,amsthm}

\usepackage[accepted]{icml2023}

\usepackage{amsmath}
\usepackage{amssymb}
\usepackage{mathtools}
\usepackage{amsthm}

\usepackage[capitalize,noabbrev]{cleveref}

\theoremstyle{plain}
\newtheorem{theorem}{Theorem}[section]
\newtheorem{proposition}[theorem]{Proposition}

\theoremstyle{definition}

\theoremstyle{remark}

\DeclareMathOperator{\sg}{sg}

\newcommand{\X}{\mathcal{X}}
\newcommand{\N}{\mathcal{N}}

\newcommand{\R}{\mathcal{R}}
\newcommand{\RR}{\mathbb{R}}
\renewcommand{\L}{\mathcal{L}}
\newcommand{\ind}{\mathbbm{1}}

\usepackage[textsize=tiny]{todonotes}

\icmltitlerunning{Supervised Metric Learning to Rank for Retrieval via Contextual Similarity Optimization}

\begin{document}

\twocolumn[
\icmltitle{Supervised Metric Learning to Rank for Retrieval via \\
           Contextual Similarity Optimization}



\icmlsetsymbol{equal}{*}

\begin{icmlauthorlist}
\icmlauthor{Christopher Liao}{bu}
\icmlauthor{Theodoros Tsiligkaridis}{mit}
\icmlauthor{Brian Kulis}{bu}
\end{icmlauthorlist}

\icmlaffiliation{bu}{Department of Electrical and Computer Engineering, Boston University}
\icmlaffiliation{mit}{MIT Lincoln Laboratory}

\icmlcorrespondingauthor{Christopher Liao}{cliao25@bu.edu}
\icmlcorrespondingauthor{Theodoros Tsiligkaridis}{ttsili@ll.mit.edu}
\icmlcorrespondingauthor{Brian Kulis}{bkulis@bu.edu}

\icmlkeywords{Machine Learning, ICML}

\vskip 0.3in
]



\printAffiliationsAndNotice{}  

\begin{abstract}
There is extensive interest in metric learning methods for image retrieval. Many metric learning loss functions focus on learning a correct ranking of training samples, but strongly overfit semantically inconsistent labels and require a large amount of data.
To address these shortcomings, we propose a new metric learning method, called \emph{contextual loss}, which optimizes \emph{contextual similarity} in addition to cosine similarity. Our contextual loss implicitly enforces \emph{semantic consistency} among neighbors while converging to the correct ranking. We empirically show that the proposed loss is more robust to label noise, and is less prone to overfitting even when a large portion of train data is withheld. Extensive experiments demonstrate that our method achieves a new state-of-the-art across four image retrieval benchmarks and multiple different evaluation settings. {\small Code is available at:  \url{https://github.com/Chris210634/metric-learning-using-contextual-similarity}}
\end{abstract}

\section{Introduction}
Image retrieval refers to learning a ranking of instances from a gallery set relative to a query image such that the highest ranked instances are the most relevant to the query. 
Several real-world applications are powered by this technology, such as person re-identification \citep{ye2021deep}, face recognition \citep{guillaumin2009you}, vehicle re-identification \citep{chu2019vehicle}, landmark retrieval \citep{weyand2020google}, and product retrieval \citep{cakir2019deep}. 
Current metric learning techniques often use a dataset with single discrete labels for supervision, and train an embedding space where images with the same label are closer together than images with different labels. However, binary supervision is unreliable, since it does not capture the complexity of relationships in the data. Furthermore, methods which overly rely on the binary supervision can be brittle in the presence of noise, since the supervision is either correct or incorrect. Multi-label datasets \citep{ranjan2015multi} mitigate this problem, but can be expensive to procure, so developing a metric learning method that is \emph{robust to label noise} and \emph{generalizable to test data} is a challenging yet important problem.

Existing image retrieval approaches fall into two main categories: classification and pairwise ranking losses. \emph{Classification losses} optimize a classifier on top of the embedding layer and discard the classifier at the end of training. \emph{Pairwise ranking losses} train the embedding layer directly by pulling together pairs of samples with the same label and pushing apart pairs of samples with different labels. Pairwise ranking methods include losses which explicitly optimize a ranking metric such as AP (average precision) surrogates: Fast-AP \cite{cakir2019deep}, Smooth-AP \cite{brown2020smooth}, Blackbox AP \cite{rolinek2020optimizing} and Roadmap \cite{ramzi2021robust}. They also include the standard contrastive, triplet and multi-similarity (MS) losses \cite{wang2019multi}. 

\begin{figure}
\centering
\includegraphics[width=1.\linewidth]{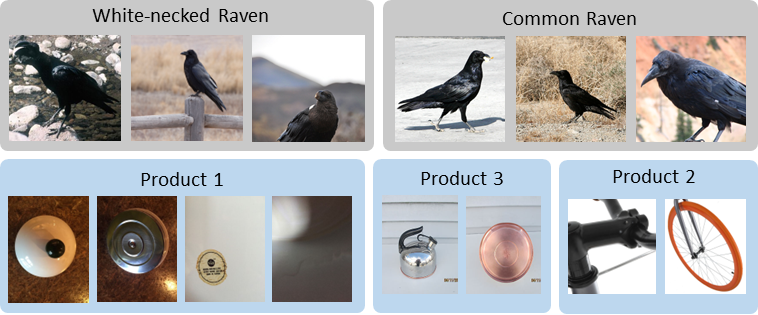}
\caption{Examples of metric learning labels which are inconsistent with semantic information from two standard benchmarks: CUB (top) and SOP (bottom). These labels are caused by a visual feature which is not present or barely visible. }
\label{fig:image_examples}
\end{figure}

\begin{figure}
\centering
\includegraphics[width=1.\linewidth]{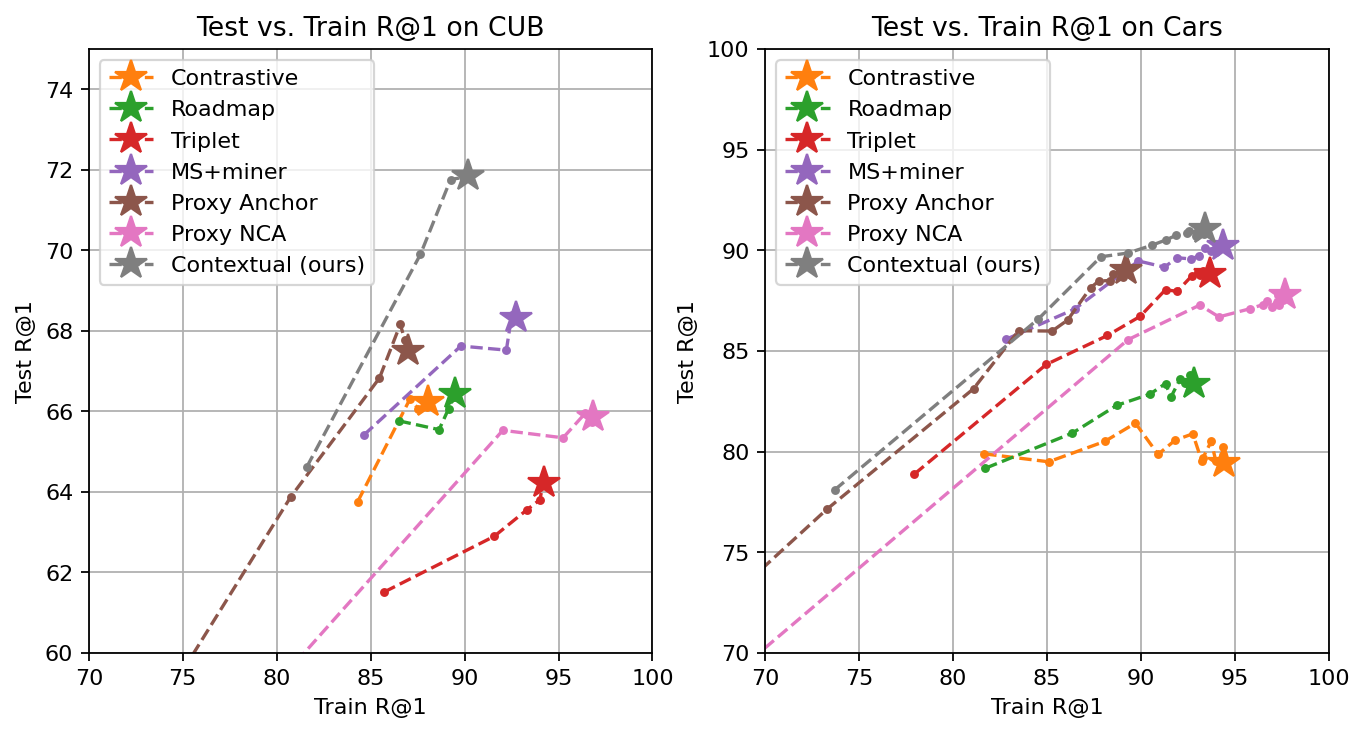}
\caption{ Comparison of our contextual loss with popular metric learning losses. We plot the test R@1 accuracy against the train R@1 accuracy over the course of training on the CUB and Cars benchmarks. The dashed line tracks the R@1 values over the course of training, and the star indicates the R@1 values at the end of training. Compared to baselines, the contextual loss achieves higher test R@1 at the expense of lower train R@1. }
\label{fig:greyedout_lambda}
\end{figure}

Empirically, classification methods, such as proxy anchor \cite{kim2020proxy}, proxy NCA \cite{teh2020proxynca++}, and HIST \cite{lim2022hypergraph} perform well on small benchmark datasets, while multi-similarity and AP surrogates perform well on large datasets. This general trend is supported by our main results in Section 5. We hypothesize that pairwise ranking methods tend to overfit the training labels while sacrificing semantic consistency of the embedding space. This can be possible even if the labels are ``correct'', as Figure \ref{fig:image_examples} illustrates. For instance, pulling apart samples of white-necked ravens from common ravens would likely lead to overfitting, since the distinguishing visual attribute is absent from the images. To address this issue, we propose to optimize \emph{contextual similarity} in addition to cosine similarity. The resulting loss function implicitly regularizes the embedding space for semantic consistency among neighbors (see results in Section 4). Figure \ref{fig:greyedout_lambda} clearly shows that our method reduces overfitting, since we achieve the best test R@1 accuracy despite lower train R@1 accuracy than some baselines. Results in Section 5 show that our method outperforms all baselines across all standard benchmarks in terms of R@1 accuracy.

\emph{Contextual similarity} is a widely used evaluation-time technique to boost retrieval accuracy. In simple terms, the contextual similarity is the fraction of neighbors two samples have in common in embedding space. Intuitively, two samples are more likely to share the same label if they have many neighbors in common, regardless of their cosine similarity. Many retrieval frameworks (\citet{zhong2017re}, \citet{cao2020unifying}) use a combination of cosine similarity and contextual similarity for evaluation, but only explicitly optimize the cosine similarity when training. In this paper, we propose to \textit{explicitly} optimize both similarities, since contextual similarity captures crucial semantic information. In another line of work, some unsupervised metric learning methods such as STML \citep{kim2022self} use contextual similarity to estimate the true similarity between unlabeled samples. Inspired by STML, we show that optimizing contextual similarity directly in the supervised setting is beneficial. As far as we know, we are the first to treat contextual similarity as a loss function for supervised learning.  This is non-trivial since contextual similarity involves non-differentiable counting operations, and as a consequence, is not amenable to off-the-shelf optimization techniques. We propose a simple but effective optimization strategy in Section 3, using heuristic gradients. We analytically justify this optimization approach in Section 4.1.

Our contributions are as follows:
\begin{enumerate}
    \item We introduce the contextual loss, which establishes a new state-of-the-art across all standard image retrieval benchmarks, even when compared to more complicated (e.g. Metrix, HIST and AVSL) and less scalable methods (AP surrogates).
    \item Our contextual loss mitigates overfitting by implicitly enforcing semantic consistency among neighbors in the embedding space. As a result, we achieve a 4\% improvement in R@1 accuracy over baselines in the presence of label noise.
    \item We conduct an extensive experimental study of our method and several popular baselines. This includes empirical results across five different benchmarks, two different experimental settings, accompanied by a comprehensive ablation study. In addition, we tune baselines extensively to promote fair comparisons.
    \item The optimization of non-differentiable steps in the loss calculation may be of interest to some readers. 
\end{enumerate}
This paper is organized as follows. Section 2 summarizes related work. Section 3 states our method, including how we optimize the non-differentiable steps in calculating contextual similarity. Section 4.1 checks that minimizing the contextual loss corresponds to learning the correct ranking of samples and that the proposed optimization procedure converges. The rest of Section 4 explores why the proposed contextual loss is \emph{less prone to overfitting} than other pairwise ranking losses by analyzing gradients and running targeted experiments. Section 5 and the Appendix present an extensive experimental study. 
{\small Code is available at:  \url{https://github.com/Chris210634/metric-learning-using-contextual-similarity}}

\section{Related Work}
\noindent \textbf{Classification Methods } We refer to any method which optimizes class centroids in conjunction with embeddings as a classification method. These methods scale with the number of classes in the training set and are usually sensitive to the learning rate of class centroids \citep{teh2020proxynca++}. Classification losses have traditionally performed well on small metric learning benchmarks; these include normalized-softmax \cite{zhai2018classification}, arcface \cite{deng2019arcface}, proxy NCA and proxy anchor. More recently, IBC \cite{seidenschwarz2021learning} and HIST \cite{lim2022hypergraph} report an improvement in R@1 when learning a graph neural network in conjunction with class centroids. However, even with these additional tricks, classification methods lag behind pairwise methods on larger benchmarks. 

\noindent \textbf{Pairwise Ranking Methods } Pairwise ranking losses include the contrastive loss \cite{hadsell2006dimensionality}, triplet loss \cite{weinberger2005distance} \cite{wu2017sampling}, multi-similarity, and AP surrogates (cited in previous section). Despite being more than a decade old,  contrastive and triplet losses  remain the go-to method for metric learning, and \citet{musgrave2020metric} show that they are comparable in performance to many recent methods. Multi-similarity includes a hard pair mining scheme that is effectively learning to rank. AP maximization methods explicitly learn to rank samples within a mini-batch. AP maximization is challenging because it involves back-propagating through the non-differentiable heaviside function, similar to the current work. As a workaround, Fast-AP uses soft-binning; Smooth-AP uses a low-temperature sigmoid; Roadmap uses an upper bound on the heaviside instead of an approximation. We find that using heuristic gradients works better for optimizing contextual similarity.

\noindent \textbf{Unsupervised Metric Learning } The concept of contextual similarity is extensively studied in the unsupervised metric learning literature, mainly in the context of person re-ID (see survey \cite{ye2021deep}). Most unsupervised person re-ID methods use the $k$-reciprocal re-rank distance \cite{zhong2017re}, which is a weighted combination of Euclidean distance and Jaccard distance between reciprocal-neighbor sets, calculated over the entire dataset. More recently, STML \cite{kim2022self} proposes an unsupervised metric learning framework for image retrieval using a simpler batch-wise contextual similarity measure. We loosely follow STML's contextual similarity definition, making significant changes to accommodate the change in problem setting and to address optimization issues (these changes are enumerated in Appendix E.1). We emphasize that prior work on contextual similarity \emph{optimizes the cosine similarity} towards the contextual similarity, focusing on the unsupervised scenario, while our work \emph{optimizes the contextual similarity} towards the true similarity, requiring full supervision.

\noindent \textbf{Robust Metric Learning } Over-reliance on binary supervision is a long-standing problem in metric learning. Many studies overcome this issue by taking advantage of the hierarchical nature of labels in metric learning datasets. \citet{sun2021dynamic}, \citet{zheng2022dynamic}, and \citet{ramzi2022hierarchical} explicitly use hierarchical labels for training. These methods assign a higher cost to mistakes in discriminating labels that are farther apart in the hierarchy, leading to a more robust embedding space. \citet{yan2021unsupervised} propose to generate synthetic hierarchical labels for unsupervised metric learning, and \citet{yan2023adaptive} extend this idea to metric learning with synthetic label noise. These two works use a hyperbolic embedding space to better capture hierarchical relationships \citep{khrulkov2020hyperbolic}. \citet{ermolov2022hyperbolic} show that simply using a hyperbolic embedding space instead of a Euclidean embedding space improves metric learning performance. Our work has a similar motivation to the above hierarchical and hyperbolic metric learning works, but we use contextual similarity instead of hierarchical labels to mitigate label inconsistency. Appendix I.4 contains some results on hierarchical retrieval metrics.

\section{Method}
\noindent \textbf{Notation } Denote the normalized output of the embedding network as $f_i \in \RR^d$. $s_{ij} = \langle f_i, f_j \rangle \in [-1,1]$ denotes the cosine similarity between the samples $i$ and $j$. There are $n$ samples in a mini-batch. We always use balanced sampling, where $k$ images are selected from $n/k$ randomly sampled labels. $n$ is divisible by $k$. $k$ is divisible by 2, but we always use $k \ge 4$ in experiments. $y_{ij} \in \{0, 1\}$ denotes the true similarity between $i$ and $j$, defined as $y_{ij}=1$ if samples $i$ and $j$ share the same label and $0$ otherwise. We use uppercase letters to denote matrices, math script to denote sets, and lowercase letters to denote scalars. $i$, $j$ and $p$ are reserved for sample indices. $\ind_{\N}$ is used to denote the binary indicator matrix for set $\N$. For instance, let $\N(i)$ denote the set of neighbors to sample $i$, then $\ind_{\N}(i,j) = 1$ if $j \in \N(i)$, and $0$ otherwise.

\noindent \textbf{Contextual Similarity Definition } 
We loosely follow the definition of contextual similarity proposed in STML \cite{kim2022self}, with significant modifications to accommodate the change in problem setting and to address optimization issues (these modifications are enumerated in Appendix E.1). 
In this section, we present the similarity definition using indicator matrices in order to show an efficient implementation in PyTorch. Note that the binary ``and'' is replaced by multiplication for differentiability. Algorithm \ref{alg:ours} contains PyTorch-like pseudo-code for Equations \ref{eq:theta_hacked} - \ref{eq:regularizer}. We include the code here for reproducibility and to show that our contextual loss can be compactly implemented despite the cumbersome mathematical notation.  

We denote the contextual similarity between samples $i$ and $j$ as $w_{ij}$. The matrix with entries $w_{ij}$ is entirely a function of the cosine similarity matrix with entries $s_{ij}$. The goal of Equations \ref{eq:theta_hacked} - \ref{eq:loss3} is to calculate $w_{ij}$ in terms of $s_{ij}$. This is implemented as {\fontfamily{qcr}\selectfont \small get\_contextual\_similarity} in Algorithm \ref{alg:ours}. For readability, we present the $w_{ij}$ calculation as three sequential steps.

\begin{algorithm}[tb]
   \caption{Pseudo-code, PyTorch-like}
   \label{alg:ours}
   \begin{minted}[fontsize=\tiny]{python}
# Hyperparameters: alpha, k, eps, s_tilde, lam, gamma
# The symbol '@' means matrix multiplication in Python
# Note: In PyTorch, set keepdim=True when calling sum(.)

class GreaterThan(autograd.Function):
    # Implements theta with heuristic gradient
    def forward(x, y):
        return (x >= y).float()
    def backward(g): # Returns gradient w.r.t (x, y)
        return g * alpha, - g * alpha
        
def get_contextual_similarity(s, k, eps):
    D = 2 - 2 * s                   # Squared Euclidean distance
    Dk = -(-D).topk(k).values[:,-1:] # Distance to k-th neighbor

    Nk_mask = GreaterThan(-D + eps, -Dk.detach())
    M_plus = (Nk_mask @ Nk_mask.T) / Nk_mask.sum(dim=1).detach()
    Nk_mask_not = 1 - Nk_mask
    M_minus = (Nk_mask_not @ Nk_mask_not.T) 
                                / Nk_mask_not.sum(dim=1).detach()
    W_1 = 0.5 * (M_plus + M_minus) * Nk_mask

    # Distance to k/2-th neighbor
    Dk_over_2 = -(-D).topk(k//2).values[:,-1:] 
    Nk_over_2_mask = GreaterThan(-D + eps, -Dk_over_2.detach())
    Rk_over_2_mask = Nk_over_2_mask * Nk_over_2_mask.T
    W_2 = (Rk_over_2_mask @ W_1) / Rk_over_2_mask.sum(dim=1)

    return 0.5 * (W_2 + W_2.T)
	
for data, labels in loader:
    f = F.normalize(model(data)) # normalized embeddings
    s = f @ f.T                  # cosine similarity matrix
    y = (labels.T == labels)     # true similarity matrix
    w = get_contextual_similarity(s, k, eps) # matrix
	
    I_neg = 1 - eye(w.shape[0])  # ones with zeros on diagonal
    L_contrast = contrastive(s, y) # Standard, code omitted
    L_reg = (s.mean() - s_tilde).square()
    L_context = ((w - s).square()* I_neg).mean()
    loss = lam * L_context + (1-lam) * L_contrast + gamma * L_reg
    loss.backward()
    optimizer.step()
    \end{minted}
\end{algorithm}

\noindent \textbf{Step 1 Neighborhood Calculation } The first step calculates a binary matrix $\ind_{\N_{k+\epsilon}}(i,j)$ indicating whether sample $j$ is a neighbor of $i$. This binary value can be thought of as a preliminary prediction of $y_{ij}$.
The neighborhood indicator calculation can be defined in terms of the heaviside function, which has no gradient. We set a constant positive gradient in the backward pass, which is reasonable since $\theta$ is a (non-strictly)
increasing function.
\begin{equation}
    \begin{split}
        &\text{Forward:  } \theta(x) = 1, \text{ if } x \ge 0 \: ; \: 0 \text{ otherwise. } \\
        &\text{Backward:  } \frac{\partial \theta(x)}{\partial x} = \alpha.
    \end{split}
\label{eq:theta_hacked}
\end{equation}
Let $D(i,j)$ denote the squared Euclidean distance between samples $i$ and $j$. By definition, $D(i,j) = 2-2s_{ij}$ and $D(i,j) \in [0,4]$. The $\sg$ operator denotes stop gradient.

Using $\theta$, we calculate the indicator function for whether sample $j$ is in the $k+\epsilon$ neighborhood of sample $i$: 
\begin{equation}
    \begin{split}
    &\ind_{\N_{k+\epsilon}}(i,j) = \theta(-D(i,j) + \sg(D(i,p)) + \epsilon),
    \\
    &\text{ where $p$ denotes the $k$-th closest neighbor of $i$.}
    \end{split}
\label{eq:loss1}
\end{equation}
In words, $\ind_{\N_{k+\epsilon}}(i,j)$ is a binary value indicating whether or not $D(i,j) \le D(i,p) + \epsilon$. By convention, the sample itself is always included in the closest neighbor count (e.g. if $k=2$, then the ``$k$-th closest neighbor'' is the closest neighbor to a sample). We now proceed to calculate the intersection of neighborhood sets.

\noindent \textbf{Step 2 Intersection of Neighborhoods } This step refines the similarity prediction by counting the number of neighbors two samples have in common. Intuitively, samples with the same label should have a similar set of neighbors. 
\begin{equation}
    \begin{split}
        &W_1(i,j) = \frac{\ind_{\N_{k+\epsilon}}(i,j)}{2} \cdot 
        \\
        & \:\:\:\:\:\:\:\:\:\:\:\: \left(\frac{M_+(i,j)}{\sg \sum_p \ind_{\N_{k+\epsilon}}(i,p) } + \frac{M_-(i,j)}{\sg \sum_p 1 - \ind_{\N_{k+\epsilon}}(i,p) } \right) ,
        \\ 
        &\text{where} \:\: M_+(i,j) = \sum_{p=1}^n \ind_{\N_{k+\epsilon}} (i, p)  \ind_{\N_{k+\epsilon}} (j, p) ,
        \\
        &\text{  and  }  M_-(i,j) = \sum_{p=1}^n \left(1 - \ind_{\N_{k+\epsilon}} (i, p) \right) \left( 1 - \ind_{\N_{k+\epsilon}} (j, p) \right) .
    \end{split}
\label{eq:intersection_of_n}
\end{equation}
$W_1(i,j) \in [0,1]$ is an intermediary similarity value. $M_+(i,j)$ counts the number of neighbors $i$ and $j$ have in common. $M_-(i,j)$ counts the number of non-neighbors $i$ and $j$ have in common. Appendix E.3 Figure \ref{fig:distribution_plots} explains why both $M_+$ and $M_-$ are necessary. The normalization factors in Eq. \ref{eq:intersection_of_n} ensure that the similarity value is between 0 and 1. We do not backpropagate gradients through the normalization factors, because it is undesirable to optimize the number of samples in the neighborhood set. As further justification for the stop gradient, note that $\sum_p \ind_{\N_{k+\epsilon}}(i,p) = k$ for any $i, p$ when $\epsilon=0$.

\noindent \textbf{Step 3 Query Expansion } This final step further refines the similarity prediction by averaging $W_1$ across close neighbors (known as query expansion, see \citet{arandjelovic2012three}).
\begin{equation}
    \begin{split}
        & \ind_{\R_{k/2 + \epsilon}} (i,j) = \ind_{\N_{k/2 + \epsilon}} (i,j) \ind_{\N_{k/2 + \epsilon}} (j,i)
        \\
        & W_2(i,j) = \frac{\sum_p \ind_{\R_{k/2 + \epsilon}} (i,p) W_1(p,j)}{ \sum_p \ind_{\R_{k/2 + \epsilon}} (i,p)}
        \\
        & w_{ij} = \frac{1}{2} \left( W_2(i,j) + W_2(j,i) \right)
    \end{split}
\label{eq:loss3}
\end{equation}
$\ind_{\R_{k/2 + \epsilon}} (i,j)$ is a binary value which equals 1 if $j$ is a $k/2+\epsilon$ neighbor of $i$ and $i$ is a $k/2+\epsilon$ neighbor of $j$, 0 otherwise. This type of reciprocal relationship is widely used in the retrieval literature, most notably by \citet{zhong2017re}. $W_2(i,j)$ is an  intermediary similarity value representing the entries of $W_1$ averaged over the smaller $\R_{k/2+\epsilon}$ neighborhood. 
$W_2$ is then symmeterized to yield the final contextual similarity values $w_{ij} \in [0,1]$. 

\noindent \textbf{Loss Function } We use the MSE loss to optimize $w_{ij}$ against the true similarity labels $y_{ij}$:
\begin{equation}
    \L_{\text{context}} = \frac{1}{n^2} \sum_{i,j|i \ne j} ( y_{ij} - w_{ij} )^2
    \label{eq:loss_context}
\end{equation}
Our final loss function $\L_{\text{ours}}$ is a sum of three loss functions:
\begin{equation}
    \L_{\text{ours}} = \lambda \L_{\text{context}} + (1 - \lambda) \L_{\text{contrast}} + \gamma \L_{\text{reg}}
    \label{eq:loss_full}
\end{equation}
\begin{equation}
      \L_{\text{reg}} =  \left( \tilde{s} - \frac{1}{n^2} \sum_{i,j}^{n^2} s_{ij} \right)^2
    \label{eq:regularizer}
\end{equation}
$\L_{\text{contrast}}$ is the standard contrastive loss (see Appendix E.3 Eq. \ref{eq:appendix_loss8}). In our work, $\L_{\text{contrast}}$ is best viewed as a regularizer that reduces the decomposability gap between the batch-wise contextual loss and the contextual loss over the entire dataset. We justify this interpretation in Appendix D Fig. \ref{fig:decomp_gap}. $\L_{\text{reg}}$ is a similarity regularizer that encourages the model to use the entire embedding space by pushing the average cosine similarity between all pairs towards the constant $\tilde{s}$.

\noindent \textbf{Remarks } The parameter $w_{ij}$ is a function of $s_{ij}$, so all three components of our loss function in Eq. \ref{eq:loss_full} optimize the cosine similarity matrix with entries $s_{ij}$. However $\L_{\text{context}}$ is the main contribution of the current work, and experiments verify that most of the improvement over baselines can be attributed to this contextual loss. The value of $k$ is not arbitrary; it must
be set to the number of samples per label in the mini-batch.
Although the time and space to calculate the contextual loss scales as $O(n^3)$, all operations are implemented as matrix multiplication, which is highly optimized on modern hardware. Appendix C Figure \ref{fig:cubic_scaling} shows that the cubic scaling is negligible for all practical batch sizes.

\noindent \textbf{Hyperparameters } $\alpha$ controls the magnitude of the heaviside gradient. Tuning $\alpha$ is unnecessary, since it is redundant with the learning rate. $\epsilon$ is the desired similarity margin between positive and negative samples. $\epsilon$ is analogous to the margin parameter on the triplet and multi-similarity loss. $\delta_{+}$ and $\delta_{-}$ (Appendix E.3 Eq. \ref{eq:appendix_loss8}) are the positive and negative margins resp. for the contrastive loss. $\tilde{s}$ is the desired average cosine similarity between all pairs. $\lambda$ and $\gamma$ control the relative weighting between the three losses. The choice of $(1 - \lambda)$ for the weight on the contrastive loss instead of a separate hyperparameter is completely arbitrary, as tuning the contrastive loss weight separately would be redundant with tuning the learning rate.

\vspace{-0.5em}
\section{Analysis}
This section discusses intuition behind the contextual loss in Eq. \ref{eq:loss_context}. Section 4.1 provides empirical evidence that $\L_{\text{context}}$ converges and shows that $\L_{\text{context}} = 0$ coincides with the correct ranking of samples. Sections 4.2 - 4.4 carefully justify the \emph{semantic consistency} argument outlined in the introduction. 

\begin{figure}
\centering
\includegraphics[width=1.\linewidth]{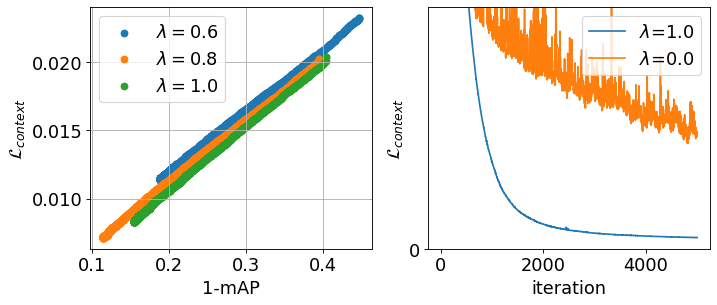}
\caption{Left: plot of contextual loss value vs. 1-mAP (mean AP) over the course of training on CUB for different choices of $\lambda$. Training proceeds from upper right to bottom left. Observe that 1-mAP decreases as contextual loss decreases. This shows that $\L_{\text{context}}$ is a valid surrogate for learning to rank. Right: convergence plot of $\L_{\text{context}}$ on CUB without mini-batching. $\L_{\text{context}}$ decreases almost monotonically when the contextual loss is minimized ($\lambda=1$), while there is a large amount of noise when the contrastive loss is minimized ($\lambda=0$). $\gamma=0$. }
\label{fig:convergence_hhh}
\end{figure}

\vspace{-0.5em}
\subsection{Contextual Loss and Optimization}
\begin{figure}
\centering
\includegraphics[width=1.\linewidth]{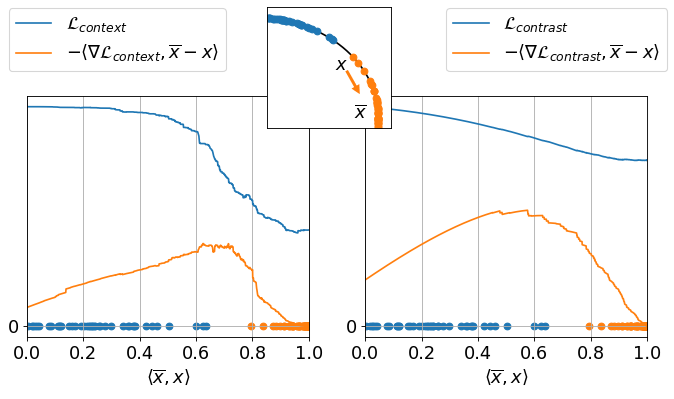}
\caption{Illustration of $\L_{\text{context}}$ and its gradient. We generate 64 random points on the unit circle from two classes, centered around the coordinates (0,1) and (1,0) (see middle plot). We move one orange point $x$ from the blue centroid to the orange centroid $\overline{x}$. The value of the loss function is plotted in blue, and the amount of gradient pointing away from $\overline{x}$ is plotted in orange. $\L_{\text{context}}$ decreases as $x$ rolls to the right, and the gradient closely approximates what it should look like if $\L_{\text{context}}$ were continuous. We include the same illustration for $\L_{\text{contrast}}$ on the right for comparison.}
\label{fig:ball_rolling}
\end{figure}

\begin{proposition}
\label{lem:usefulproposition}
For a batch of size $n$ with exactly $k$ samples from each class ($n$ divisible by $k$, $n > 2k$, and $k \ge 2$), assuming that $\epsilon=0$, $\L_{\text{context}} = 0$ if and only if all samples are correctly ranked with respect to every other sample within the batch, i.e. $ s_{ip} > s_{ij} ,\:\forall p,j \text{ where } y_{ij} = 0 \text{ and } y_{ip} = 1 ,\:\forall i \in [1,n]$.
\end{proposition}
We defer the proof to Appendix B. This Proposition shows that $\L_{\text{context}}$ is a valid ranking objective, similar to AP surrogates, multi-similarity, and triplet losses. Note that Proposition \ref{lem:usefulproposition} does not hold for $\L_{\text{contrast}}$, since $\L_{\text{contrast}}$ continues to provide gradients up to fixed margins, regardless of whether the correct ranking is satisfied. Figure \ref{fig:convergence_hhh} (left) shows that $\L_{\text{context}}$ is approximately a linearly scaled version of 1-mAP over the course of training. Figure \ref{fig:convergence_hhh} (right) suggests that the value of $\L_{\text{context}}$ converges when optimized using gradient descent. The Appendix contains more empirical evidence that the value of $\L_{\text{context}}$ converges (Fig. \ref{fig:decomp_gap} and \ref{fig:loss_plot} ). Figure \ref{fig:ball_rolling} justifies the choice of heuristic gradient in Eq. \ref{eq:theta_hacked}. In this simple 2-D example, the gradient is always positive and non-zero in the direction away from the minimum, until the minimum is reached.

\vspace{-0.5em}
\subsection{Intuition}
In the previous subsection, we proved that the minimum of $\L_{\text{context}}$ corresponds to a correct ranking of samples within a batch. We also showed that the value of $\L_{\text{context}}$ converges empirically. However, we still need some intuition as to why gradients from $\L_{\text{context}}$ work better than simple pair-wise contrastive loss functions. This discussion will naturally lead to the \emph{semantic consistency} intuition promised at the beginning of the paper. Let us start by asking: \emph{what is the value of optimizing the intersection of neighborhood sets in the manner of Eq. \ref{eq:intersection_of_n}?} We offer a straight-forward intuition: \emph{Maximizing the intersection between the neighborhood sets of two samples is equivalent to pushing one sample towards the context of the other sample and vice versa}; \emph{minimizing the intersection is equivalent to pulling apart the contexts of the two samples}. We show this intuition by analyzing the gradient.

\begin{figure}
\centering
\includegraphics[width=1.\linewidth]{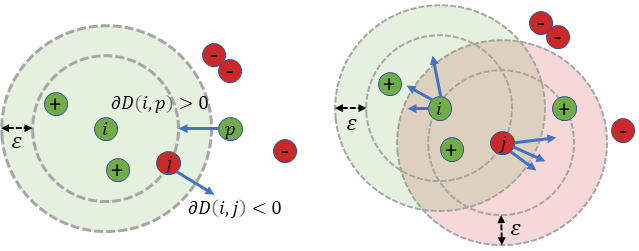}
\caption{This figure complements Eq. \ref{eq:ana4}. Part ${\large \textcircled{\small 1}}$ of the gradient (left) enforces correct ranking w.r.t. sample $i$. Part ${\large \textcircled{\small 2}}$ of the gradient (right) increases the distance between $i$ and all samples in the neighborhood of $j$, and vice versa. As shown, sometimes this implies that samples with the same label are pulled apart. The $k+\epsilon$ neighborhood as defined in Eq. \ref{eq:loss1} is shaded. $k=4$.}
\label{fig:target_picture_inter}
\end{figure}

For simplicity, consider $\epsilon = 0$, such that the normalization factors in Eq. \ref{eq:intersection_of_n} are constant: $ \sum_p \ind_{\N_{k+\epsilon}}(i,p) = k $ and  $\sum_p 1 - \ind_{\N_{k+\epsilon}}(i,p) = n-k $. For the remainder of the section we drop the $k+\epsilon$ subscript from $\N$ for readability. Further consider a negative pair of samples $i$ and $j$  ($y_{ij} = 0$), where $j$ is wrongly ranked w.r.t. $i$ (i.e. $\ind_{\N} (i, j) = 1$). For the sake of developing intuition, let us further assume that all entries of $W_1$ are correct except index $i,j$; also, $\ind_{\R_{k/2 + \epsilon}} (i,p) = 0 \: \forall p \ne i$. Under these assumptions, $w_{ij} = W_1(i,j)$ and $\L_{\text{context}} = \frac{1}{2n^2} (y_{ij} - w_{ij})^2$. This allows us to focus on interpreting Eq. \ref{eq:intersection_of_n}. Under these assumptions, Eq. \ref{eq:intersection_of_n} simplifies to (see algebra in Appendix A):
\begin{equation}
    \begin{split}
        & W_1(i,j) = \underbrace{\ind_{\N} (i, j)}_{{\large \textcircled{\small 1}} } \underbrace{\left( a \langle \ind_{\N}(i), \ind_{\N}(j) \rangle + b \right)}_{ {\large \textcircled{\small 2}} },
        \\
        & \text{ where } a = \frac{1}{2k} + \frac{1}{2(n-k)} , \text{ and } b = \frac{n-2k}{2(n-k)}.
    \end{split}
    \label{eq:ana3}
\end{equation}
$a$ and $b$ are positive constants, assuming that $n > 2k$. $\langle \cdot \rangle$ denotes the inner product between the $i$-th and $j$-th rows of the indicator matrix. This inner product is clearly positive. $W_1 (i,j) \in (0,1]$ must be a non-zero positive number under our assumptions. The gradient w.r.t. $W_1 (i,j)$ must be non-zero positive because $y_{ij} = 0$. We are now ready for the backward pass. For simplicity of notation, $\partial W_1 (i,j) := g_{ij} > 0$ denotes the gradient of the loss w.r.t. $W_1 (i,j)$. The gradient w.r.t. the distance matrix $\partial D$ can be split into two parts ${\large \textcircled{\small 1}}$ and ${\large \textcircled{\small 2}}$, added together by chain rule.
\begin{equation}
\begin{split}
&{\large \textcircled{\small 1}} \:
\partial D(i,j) = -\alpha \partial \ind_{\N} (i, j) = -\alpha g_{ij} \left( a \langle \cdot \rangle + b \right)
\\
&{\large \textcircled{\small 2}}
    \begin{cases}
      \partial D(i) = -\alpha \partial \ind_{\N} (i) &=  -\alpha ag_{ij} \ind_{\N} (i, j) \ind_{\N} (j) \\
      \partial D(j) = -\alpha \partial \ind_{\N} (j) &=  -\alpha ag_{ij} \ind_{\N} (i, j) \ind_{\N} (i) \\
    \end{cases}
\end{split}
\label{eq:ana4}
\end{equation}
Note that the $-\alpha$ factors in Eq. \ref{eq:ana4} come from going backward through $\theta(\cdot)$. The negative sign accounts for optimizing distance instead of similarity. Intuitively, the two components of the gradient perform different functions. The gradient in Eq. \ref{eq:ana4} ${\large \textcircled{\small 1}}$ enforces correct ranking. The gradient in Eq. \ref{eq:ana4} ${\large \textcircled{\small 2}}$ pulls $i$ away from the context of $j$ and $j$ away from the context of $i$. More clearly, $\partial D(i,p) < 0$ when $\ind_{\N} (j,p) = 1$ and $\partial D(i,p) = 0$ otherwise, for all samples $p$ in the batch. In words, we increase the distance between $i$ and all samples in \emph{the neighborhood} of $j$. See Figure \ref{fig:target_picture_inter} for an illustration.

\subsection{Semantic Consistency}
The previous subsection showed that the gradients of the contextual loss optimize distances between neighbors of samples, not just pairwise distances. This is important because the neighborhood of a sample contains semantically similar images, regardless of whether they have the same label or not. Indeed, optimizing contextual similarity may result in pulling apart samples with the \emph{same label}, and to a lesser extent pushing together samples with \emph{different labels}. For example, in Figure \ref{fig:target_picture_inter}, sample $i$ is pulled both from sample $j$, which has a different label, and from the two neighbors of sample $j$, which have the same labels as sample $i$. 

This behavior is novel to the contextual loss. All of our baselines which take the cosine similarity matrix as input satisfy the condition that $\partial D(i,j) \ge 0 $ when $y_{ij} = 1$ and $\partial D(i,j) \le 0 $ when $y_{ij} = 0$. Note that this discussion is limited to pairwise ranking losses; classification methods cannot be analyzed in this way. We analyze the effect of these ``wrong'' gradients in Table \ref{tab:smalltable} by adding a custom auto-grad function between the distance matrix and the contextual loss. This function truncates the gradient of the distance matrix according to the label matrix:
\begin{equation}
\partial D(i,j) = 
    \begin{cases}
      \max(\partial D(i,j), 0), & \text{if } y_{ij} = 1 \\
      \min(\partial D(i,j), 0), & \text{otherwise}.
    \end{cases}
\label{eq:tuncategrads}
\end{equation}
Clearly, clamping the distance gradients according to Eq. \ref{eq:tuncategrads} raises the R@1 on training data at the expense of a lower test R@1. This raises the question: \emph{why does discarding seemingly wrong gradients lower the R@1 accuracy by 6\%?} We hypothesize that $\L_{\text{context}}$ implicitly regularizes the embedding space against the label and image noise illustrated in Fig. \ref{fig:image_examples}. The contextual loss considers relationships between groups of $k$ samples, which intuitively should be more robust to random label variations than solely relying on pairwise relationships. In other words, sample pairs with gradients $\partial D(i,j)$ that violate the true labels $y_{ij}$ are pairs where the $y_{ij}$ is inconsistent with semantic information; these labels likely do not generalize to test data.  We justify this hypothesis in the next section by testing the robustness and generalizability of our approach.

\begin{table}
\caption{Comparison of train and test accuracy on CUB between $\L_{\text{context}}$ and $\L_{\text{context}}$ with gradient corrected according to Eq. \ref{eq:tuncategrads}.}
\centering
{
\begin{tabular}{l c c}
\toprule
 CUB & $\L_{\text{context}}$ & with gradient correction\\
\midrule
Train R@1 & 87.0 & 92.9 \\
Test R@1 & 71.4 & 65.4 \\
\bottomrule
\end{tabular}
}
\label{tab:smalltable}
\end{table}

\begin{figure}
\centering
\includegraphics[width=1.\linewidth]{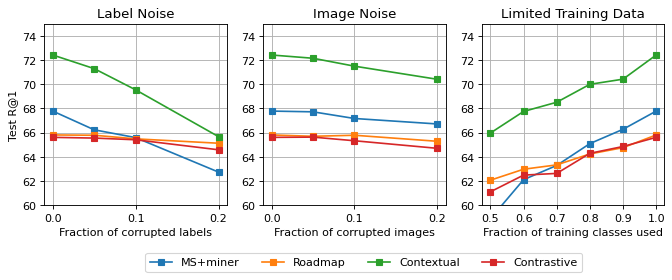}
\caption{Robustness and generalizability comparison of the contextual loss against other pairwise ranking losses. See description in Section 4.4. For this experiment, we test $\L_{\text{context}}$ without additional regularization, i.e. $\lambda=1$ and $\gamma=0$. }
\label{fig:robustness}
\end{figure}

\vspace{-0.5em}
\subsection{Robustness and Generalizability Experiments}
The previous subsection proposed that optimizing contextual similarity with $\L_{\text{context}}$ leads to an embedding space with higher test R@1 accuracy because contextual similarity is more robust to label noise and more generalizable to test data. We verify this claim with three sets of experiments on the CUB benchmark in Figure \ref{fig:robustness}.

\vspace{-0.3em}
\noindent \textbf{Label Noise } We experiment with assigning random labels to 5, 10, and 20 \% of randomly selected training samples. 

\vspace{-0.3em}
\noindent \textbf{Image Noise } We scraped around 1,800 generic bird images from Bing using search terms such as ``bird'' and ``flying bird''. We manually filtered the images such that each image contains at least one bird. We then replace a percentage of randomly selected CUB training samples with random images from our Bing dataset. This experiment simulates a typical web-scraped dataset, where images are often only lightly proofread by a human. In most modern image datasets, a small portion of labels are either wrong or do not accurately reflect the content of the image. We experiment with 5, 10, and 20 \% image noise. 

\vspace{-0.3em}
\noindent \textbf{Limited Training Data } According to traditional machine learning wisdom, small datasets are easier to overfit. We demonstrate that our method achieves reasonable results even when some training data is withheld. CUB-200 is already the smallest standard benchmark, with 5,994 training samples belonging to 100 classes. We experiment with using only the first 50, 60, 70, 80, and 90 classes for training. 

\vspace{-0.3em}
The results from these three sets of experiments are presented in Fig. \ref{fig:robustness}. Our method clearly outperforms other pairwise ranking losses. We tune learning rates individually for each method. We use batch size 256 and $k=8$. We use $\L_{\text{context}}$ by itself instead of the entire loss to study the contextual loss without additional regularization.

\begin{table*}
\caption{ State-of-the-art comparisons on CUB and Cars datasets. We use ResNet-50. The last two rows use embedding size 1536, while the other rows use embedding size 512. Results should not be compared across embedding sizes. $\dagger$ indicates our reproduction using the implementation by \citet{musgrave2020pytorch}; other results are copied from their original paper. All results indicated by $\dagger$ are an average of 6 trials with different random seeds but with the same train-test split. }
\centering
{  \small
\begin{tabular}{ l >{\columncolor[gray]{0.9}}c c c c c >{\columncolor[gray]{0.9}} c c c c c  }
 \toprule
 & \multicolumn{5}{c}{CUB} & \multicolumn{5}{c}{Cars}  \\
 \midrule
Method & R@1 & R@2 & R@4 & R@8 & mAP & R@1 & R@2 & R@4 & R@8 & mAP\\
 \midrule
DRML \cite{zheng2021deep} & 68.7 & 78.6 & 86.3 & 91.6 &  - & 86.9 & 92.1 & 95.2 & 97.4 & -\\
DIML \cite{zhao2021towards}   & 68.2 & - & - & - &  - & 87.0 & - & - & - & -\\
DiVA \cite{milbich2020diva}   & 69.2 & 79.3 & - & -  & - & 87.6 & 92.9 & - & - &-\\
Proxy Anchor \cite{kim2020proxy}   & 69.7 & 80.0 & 87.0 & 92.4 &  - & 87.7 & 92.9 & 95.8 & 97.9 &-\\
MS \cite{wang2019multi}   & 67.8 & 77.8 & 85.6 &  - &  - & 87.8 & 92.7 & 95.3 &  - &-\\
IBC \cite{seidenschwarz2021learning}   & 70.3 & 80.3 & 87.6 &  - &  - & 88.1 & 93.3 & 96.2 &  - &-\\
S2SD \cite{roth2020s2sd}   & 70.1 & 79.7 &  - &  - &  - & 89.5 & 93.9 &  - &  - &-\\
Proxy NCA + Metrix   & 70.4 & 80.6 & \textbf{88.7} &  - &  - & 88.5 & 93.4 & 96.5 &  - &-\\
PA + Metrix   & 71.0 & \textbf{81.8} & 88.2 &  - &  - & 89.1 & 93.6 & 96.7 &  - &-\\
MS + Metrix \cite{venkataramanan2021takes}   & 71.4 & 80.6 & 86.8 &  - &  - & 89.6 & 94.2 & 96.0 &  - &-\\
HIST \cite{lim2022hypergraph}   & 71.4 & 81.1 & 88.1 &  - &  - & 89.6 & 93.9 & 96.4 &  - &-\\
MHGL \cite{ebrahimpour2022multi}   & 70.6 & 80.9 & 88.0 & 92.3 &  - & 90.1 & 94.2 & 96.4 & 98.1 &-\\
MS $\dagger$ \cite{wang2019multi}  & 65.9 & 76.6 & 84.9 & 90.8 & 36.1 & 82.1 & 88.6 & 92.9 & 95.7 & 35.7\\
Contrastive $\dagger$ \cite{hadsell2006dimensionality}  & 65.9 & 76.6 & 84.8 & 90.8 & 35.3 & 82.4 & 88.7 & 92.8 & 95.6 & 35.3\\
Roadmap $\dagger$ \cite{ramzi2021robust}  & 66.0 & 76.8 & 85.3 & 91.1 & 36.0 & 83.5 & 89.8 & 93.7 & 96.3 & 37.3\\
Triplet $\dagger$ \cite{weinberger2005distance}  & 64.8 & 75.9 & 84.5 & 90.6 & 33.8 & 88.1 & 93.1 & 96.0 & 97.7 & 41.8\\
MS + miner $\dagger$ \cite{wang2019multi}  & 68.0 & 78.4 & 86.2 & 91.7 & 36.4 & 90.5 & 94.7 & 97.0 & \textbf{98.4} & 42.7\\
Proxy Anchor $\dagger$ \cite{kim2020proxy}  & 69.1 & 79.5 & 87.0 & 92.2 & 37.4 & 89.0 & 93.5 & 96.2 & 97.9 & 38.8\\
Proxy NCA $\dagger$  \cite{teh2020proxynca++} & 66.3 & 77.1 & 85.5 & 91.5 & 35.1 & 88.2 & 93.2 & 96.1 & 97.9 & 38.2\\
\textbf{Contextual (Ours)} $\dagger$   & \textbf{71.9} & 81.5 & 88.5 & \textbf{93.1} & \textbf{40.2} & \textbf{91.1} & \textbf{95.0} & \textbf{97.1} & \textbf{98.4} & \textbf{43.3}\\
 \midrule
PA + AVSL \cite{zhang2022attributable} & 71.9 & 81.7 & 88.1 & 93.2 &  - & 91.5 & 95.0 & 97.0 & 98.4 &-\\
\textbf{Contextual (Ours)} $\dagger$ & \textbf{72.7} & \textbf{82.2} & \textbf{88.8} & \textbf{93.4} & \textbf{40.5} & \textbf{91.8} & \textbf{95.4} & \textbf{97.4} & \textbf{98.6} & \textbf{43.3} \\
 \bottomrule
\end{tabular}
}
\label{tab:main1}
\end{table*}
\begin{table}
\caption{ State-of-the-art comparisons on SOP. We use ResNet-50. The last two rows use embedding size 1536, while the other rows use 512. All results indicated by $\dagger$ are an average of 2 trials with different random seeds but with the same train-test split. }
\centering
{  \small
\begin{tabular}{ l  >{\columncolor[gray]{0.9}}c c c c  }
 \toprule
Method & R@1 & R@10 & R@100 & R@1000\\
 \midrule
DRML & 79.9 & 90.7 & 96.1 & -\\
DIML & 79.3 & - & - &-\\
DiVA & 79.6 & 91.2 & - &-\\
Proxy Anchor & 79.1 & 90.8 & 96.2 &-\\
MS & 76.9 & 89.8 & 95.9 &-\\
IBC & 81.4 & 91.3 & 95.9 &-\\
S2SD & 80.0 & 91.4 & - &-\\
Proxy NCA + Metrix & 81.3 & \textbf{92.7} & 97.1 &-\\
PA + Metrix & 81.3 & 91.7 & 96.9 &-\\
MS + Metrix & 81.0 & 92.0 & \textbf{97.2} &-\\
HIST & 81.4 & 92.0 & 96.7 &-\\
MHGL & 81.7 & 92.0 & 96.6 &-\\
MS $\dagger$ & 79.9 & 90.4 & 95.8 & 98.6\\
Contrastive $\dagger$ & 80.9 & 90.9 & 95.6 & 98.3\\
Roadmap $\dagger$ & 81.9 & 92.0 & 96.3 & 98.7\\
Triplet $\dagger$ & 81.9 & 92.5 & 96.8 & \textbf{98.9}\\
MS + miner $\dagger$ & 82.2 & 92.5 & 96.7 & 98.8\\
Proxy Anchor $\dagger$ & 79.7 & 91.1 & 96.2 & 98.7\\
Proxy NCA $\dagger$ & 78.8 & 90.7 & 96.3 & 98.8\\
\textbf{Contextual (Ours)} $\dagger$ & \textbf{82.6} & 92.5 & 96.7 & 98.8\\
\midrule
PA + AVSL & 79.6 & 91.4 & 96.4 & - \\
\textbf{Contextual (Ours)} $\dagger$ & \textbf{83.2} & \textbf{92.9} & \textbf{96.8} & \textbf{98.8} \\
 \bottomrule
\end{tabular}
}
\label{tab:main2}
\end{table}

\begin{figure}
\centering
\includegraphics[width=1.\linewidth]{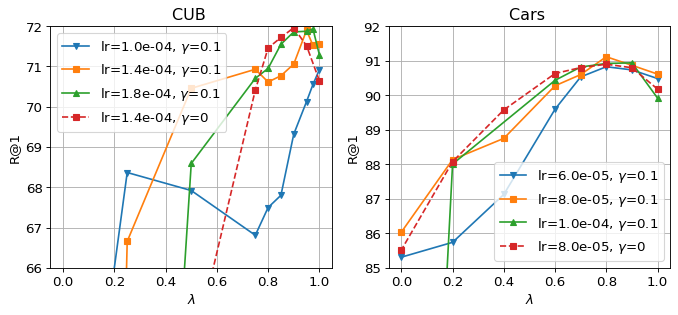}
\caption{Ablation results. We test the contribution of each component in Eq. \ref{eq:loss_full} by trying different values for $\lambda$ and $\gamma$ on CUB and Cars. The dashed line indicates results without the similarity regularizer ($\gamma = 0$). The three different marker symbols represent different learning rates. We show results for different learning rates because the optimal learning rate varies with $\lambda$. Observe that the optimal R@1 is always achieved by a combination of $\L_{\text{contrast}}$ and $\L_{\text{context}}$. The similarity regularizer is sometimes helpful, but never detrimental.}
\label{fig:ablation}
\end{figure}

\vspace{-0.5em}
\section{Experiments}
\noindent \textbf{Datasets } We experiment on two small-scale and two large-scale datasets: Caltech-UCSD Birds (CUB-200) \cite{wah2011caltech}, Stanford Cars-196 \cite{krause20133d}, Stanford Online Products (SOP) \cite{oh2016deep}, and mini-iNaturalist-2021 \cite{van2018inaturalist}. CUB-200 and Cars-196 are smaller fine-grain classification datasets with 200 and 196 unique labels, respectively. SOP is a large-scale dataset with 120,053 product images from 22,634 classes. mini-iNaturalist-2021 is a subset of the iNaturalist-2021 species classification competition dataset, with 50 images from each of 10,000 species. iNaturalist results are deferred to Appendix G Table \ref{tab:main}.

\noindent \textbf{Baselines } We compare against a diverse set of baselines in Tables  \ref{tab:main1} and \ref{tab:main2}.  Some baselines (such as DRML, Metrix, S2SD, HIST, MHGL, and AVSL) use complicated tricks to achieve published results. We simply copy the results for these baselines from their original paper. We then choose a representative set of baselines which only modify the loss function, and reproduce their results under identical experimental conditions (indicated by $\dagger$ in the tables). We first compare against contrastive and triplet losses, which are the accepted standard in the field. Multi-similarity (MS) is a popular pairwise ranking loss. From classification methods, we compare against proxy anchor and proxy NCA. From AP maximization methods, we compare against Fast-AP, Smooth-AP, and Roadmap (Fast-AP and Smooth-AP comparisons are deferred to Appendix G Table \ref{tab:main}). The benchmark results for many of the above-mentioned loss functions are under-represented in the literature. For fair comparison, we tune learning rates separately for each loss function. We use default values for any other hyperparameters.

\noindent \textbf{Hyperparameters and Setup } On our method, we tune $\lambda$ and $\epsilon$ separately for each dataset. We use fixed values for remaining hyperparameters: $\gamma=0.1$, $\alpha=10.0$, $k=4$, $\delta_+ = 0.75$, $\delta_- = 0.6$, $\tilde{s} = 0.3$. The results in the main paper all use 224$\times$224 image resolution. Some recent studies use 256$\times$256 image resolution; comparisons in this setting are included in Appendix G Table \ref{tab:main}. We use Adam with a decaying learning-rate schedule. We report results on the model with the best test R@1 metric, as is standard in the literature. We tune learning rates separately for each method and dataset combination. We use a batch size of 256 for iNaturalist, 128 for SOP and CUB, and 64 for Cars; the larger batch size is necessary to achieve reasonable performance on iNaturalist, while the smaller batch size appears to reduce overfitting on Cars. We use a 4 per class balanced sampler. For SOP and iNaturalist, we use hierarchical sampling \citep{cakir2019deep}, following prior work. We use an embedding size of 512 for most comparisons; we only use an embedding size of 1536 for comparison with AVSL. We use ResNet-50 with a linear embedding layer. For CUB and Cars, we add an additional linear projector layer, which is discarded at the end of training. We use GeM pooling \cite{radenovic2018fine}, a widely used generalization of max and mean pooling. We always freeze batch-norm. 
We emphasize that the setup described above is used for all baseline results marked by $\dagger$. Additionally, we add the linear projector and/or similarity regularization with $\gamma=0.1$ only when it improves the baseline, for fair comparison. 

\noindent \textbf{Performance Metrics } We report Recall @ $k$ for select $k$. R@$k$ is the percentage of test samples where at least one of the $k$ closest neighbors have the same label. We also report mAP, a standard ranking metric defined in Appendix I.3. We report the average of 6 trials on CUB and Cars, and 2 trials on SOP. We omit standard deviations for readability. 

\noindent \textbf{Discussion } Our R@1 results are better than the best baseline across all datasets and embedding sizes. We achieve R@1 gains of 0.5\%, 0.6 \%, and 0.4 \% on CUB, Cars, and SOP resp. for the 512 embedding size. We achieve gains of 0.8\%, 0.3\%, and 3.6\% on these datasets for the 1536 embedding size. These results are an average of 6 trials with different random seeds on CUB and Cars, and 2 trials on SOP.
Additionally, we achieve R@1 gains of 0.8\%, 0.6\%, 0.2\%, and 0.3\% over the best baseline on CUB, Cars, SOP, and iNaturalist, resp. averaged over 3 trials, under slightly different experimental settings using 256$\times$256 image resolution (see Table \ref{tab:main} in the appendix).

\noindent \textbf{Ablation } Figure \ref{fig:ablation} plots R@1 on CUB and Cars with different values of $\lambda$ and $\gamma$. This figure shows that the best R@1 performance is always achieved by a combination of contextual and contrastive losses. The optimal value of $\lambda$ is usually between 0.8 and 0.9. The similarity regularizer is only needed to achieve state-of-the-art on some datasets. This property is reasonable for a regularizer, and it is more important to observe that the similarity regularizer improves the results of many diverse metric learning losses (see Table \ref{tab:regsim-results} Appendix I.1).

\vspace{-0.5em}
\section{Conclusion}
In this work, we proposed a novel contextual loss based on contextual similarity optimization. Our contextual loss improves the R@1 performance significantly over the current state-of-the-art across four benchmarks, when regularized by the contrastive loss and a novel similarity regularizer. We established that our loss function reduces overfitting by regularizing the embedding space for \emph{semantic consistency among neighbors}. We justified this interpretation both analytically by inspecting the gradient, and empirically by showing that our loss function is more robust to label noise. We carefully supported each component of our loss function through extensive experiments across two different experimental settings, accompanied by exhaustive ablation studies.

\subsubsection*{Ethics Statement}
We note that metric learning can be applied to controversial problems such as person re-identification and face re-identification. Our work is mainly foundational, so does not contribute directly to these applications. We also limit our experimentation to the image retrieval aspect of metric learning.

\subsubsection*{Reproducibility Statement}
Instructions on how to run the code is provided in a README file. We include details on hardware requirements in Appendix H.1.
Code is released here: \url{https://github.com/Chris210634/metric-learning-using-contextual-similarity}

\subsubsection*{Acknowledgments}
DISTRIBUTION STATEMENT A. Approved for public release. Distribution is unlimited. This material is based upon work supported by the Under Secretary of Defense for Research and Engineering under Air Force Contract No. FA8702-15-D-0001. Any opinions, findings, conclusions or recommendations expressed in this material are those of the author(s) and do not necessarily reflect the views of the Under Secretary of Defense for Research and Engineering.

\bibliography{example_paper}
\bibliographystyle{icml2023}

\newpage
\appendix
\onecolumn
\section{Simplification of Step 2 in Main Paper Section 3}
Under the assumption that $\epsilon = 0$, the formula for $W_1$ in terms of $\ind_{\N_k}$ simplifies. We drop the subscript $k$ from $\N$ for readability. When $\epsilon = 0$, there are exactly $k$ samples in the neighborhood set $\N$. Specifically, $\sum_p \ind_{\N}(i,p) = k$ and $\sum_p 1 - \ind_{\N}(i,p) = n-k$. Eq. \ref{eq:intersection_of_n} becomes:
\begin{equation}
    \begin{split}
        &W_1(i,j) = \frac{\ind_{\N}(i,j)}{2} \cdot \left(\frac{M_+(i,j)}{k } + \frac{M_-(i,j)}{n - k } \right) ,
        \\ 
        &\text{where} \:\: M_+(i,j) = \sum_{p=1}^n \ind_{\N} (i, p)  \ind_{\N} (j, p) ,
        \\
        &\text{  and  }  M_-(i,j) = \sum_{p=1}^n \left(1 - \ind_{\N} (i, p) \right) \left( 1 - \ind_{\N} (j, p) \right) .
    \end{split}
\label{eq:appendix_loss2_sim_1}
\end{equation}
Under our assumptions, $M_-(i,j)$ simplifies:
\begin{equation}
\begin{split}
    M_-(i,j) & = \sum_{p=1}^n \left( 1 - \ind_{\N} (i, p) - \ind_{\N} (j, p) + \ind_{\N} (i, p)\ind_{\N} (j, p)\right)
    \\
    &= n - k - k + \sum_{p=1}^n\ind_{\N} (i, p)\ind_{\N} (j, p)
    \\
    &= n - 2k + \langle \ind_{\N}(i), \ind_{\N}(j) \rangle
\end{split}
\end{equation}
Note that  $M_+(i,j) = \langle \ind_{\N}(i), \ind_{\N}(j) \rangle$, by definition. Eq. \ref{eq:appendix_loss2_sim_1} becomes:
\begin{equation}
    W_1(i,j) = \frac{\ind_{\N}(i,j)}{2} \cdot \left(\frac{\langle \ind_{\N}(i), \ind_{\N}(j) \rangle}{k } + \frac{n - 2k + \langle \ind_{\N}(i), \ind_{\N}(j) \rangle}{n - k } \right) 
\label{eq:appendix_w_1_app}
\end{equation}
This can be written as Eq. \ref{eq:ana3} in the main paper, restated here:
\begin{equation}
         W_1(i,j) = \ind_{\N} (i, j) \left( a \langle \ind_{\N}(i), \ind_{\N}(j) \rangle + b \right), \text{ where } a = \frac{1}{2k} + \frac{1}{2(n-k)} \text{ and } b = \frac{n-2k}{2(n-k)}
\label{eq:appendix_w_2_app}
\end{equation}
Under the assumption that $n > 2k$, $a>0$ and $b>0$. 

\section{Proof of Proposition \ref{lem:usefulproposition}}
\noindent \textbf{Forward direction } If all samples are correctly ranked w.r.t. every other sample within the batch, show that $\L_{\text{context}} = 0$.

\emph{Proof } Recall that the Proposition assumes that there are exactly $k$ samples from each class in the batch. This means that the $k$ neighborhood set $\N(i)$ of every sample $i$ contains exactly the $k$ samples in the batch with the same label. In other words, $\ind_{\N}(i,j) = y_{ij}, \: \forall i,j$. Further, if $\ind_{\N}(i,j) = 1$, then the inner product $\langle \ind_{\N}(i), \ind_{\N}(j) \rangle = k$ because the neighborhood sets of $i$ and $j$ are identical and there are exactly $k$ samples in this set. 

When $\langle \ind_{\N}(i), \ind_{\N}(j) \rangle = k$,  $W_1(i,j) = \ind_{\N} (i, j) $. When $\ind_{\N}(i,j) = 0$, $W_1(i,j) = 0$. Therefore, $W_1(i,j) = \ind_{\N} (i, j) = y_{ij}, \: \forall i,j$.

Following Eq. \ref{eq:loss3}, it is also clear that $w_{ij} = W_1(i,j)$ when samples are correctly ranked. This is because, the $\R_{k/2}$ neighborhood is a subset of the larger $\N_{k}$ neighborhood. That is, if $\ind_{\R_{k/2}}(i,j) = 1$, then $\ind_{\N}(i,j) = 1$. Since $\ind_{\N}(i,p) = \ind_{\N}(j,p), \:\forall p$, when $\ind_{\N}(i,j) = 1$, averaging the entries of $W_1$ over $\R_{k/2}$ as indicated in Eq. \ref{eq:loss3} does not change $W_1$. Therefore, $W_2(i,j) = W_1(i,j)$. Finally, $y_{ij}$ is symmetric, so $W_2(i,j) = w_{ij}$. This proves that $y_{ij} = w_{ij}$ and $\L_{\text{context}} = 0$ by Eq. \ref{eq:loss_context}. 

\noindent \textbf{Backward direction } If $\L_{\text{context}} = 0$, show that all samples are correctly ranked w.r.t. every other samples within the batch.

\emph{Proof } We prove this by contradiction. Suppose that $\L_{\text{context}} = 0$ and there exists a pair of samples $i$, $j$ where $j$ is not correctly ranked w.r.t. $i$. There are two possible cases: (1) $y_{ij} = 1$ but $\ind_{\N}(i,j) = 0$ or (2) $y_{ij} = 0$ but $\ind_{\N}(i,j) = 1$. (1) is the case where $i$ and $j$ have the same label, but $j$ is not ranked as one of the top-$k$ neighbors of $i$. (2) is the case where $i$ and $j$ have different labels, but $j$ is ranked as one of the top-$k$ neighbors of $i$. Since we assume that there are exactly $k$ samples with the same label per batch, both (1) and (2) must be true. In particular, assume that (2) is true. Under Eq. \ref{eq:appendix_w_2_app}, if $\ind_{\N}(i,j) = 1$, then $W_1(i,j) > 0$ because $\ind_{\N}(i,j)$ is multiplied by a non-zero positive number. Similarly, under Eq. \ref{eq:loss3}, if $W_1(i,j) > 0$, then $w_{ij} > 0$ because $W_1(i,j)$ is multiplied by non-zero positive numbers and added to positive numbers. Since $w_{ij} > 0$ and $y_{ij}=0$, $\L_{\text{context}} > 0$, which contradicts the assumption that $\L_{\text{context}} = 0$ and concludes the proof.

\begin{figure}
\centering
\includegraphics[width=0.8\linewidth]{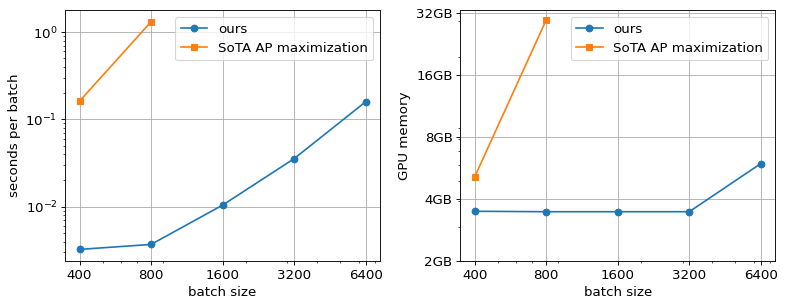}
\caption{Comparison of scalability in time and space between our loss function and Roadmap. Our method is plotted in blue, while Roadmap is plotted in orange. For this experiment, we pre-calculate the 2048-dimensional ResNet-50 features and train a linear embedding layer using each loss function. Both the time taken per batch (left) and GPU memory used (right) scale cubicly. However, the time and space taken to calculate our loss function is negligible compared to backbone evaluation, even for batch size 6400. On the other hand, the time and space needed to calculate the Roadmap loss quickly becomes unpractical for large batch sizes. Note: we pre-compute all features and store them in GPU memory. The memory used by the pre-computation is freed, but the GPU memory being used, as indicated by the Nvidia tool, remains at the maximum memory used over the lifetime of the process. Therefore, the memory consumed by calculating the loss function is not visible until it exceeds the memory used by the pre-computation. }
\label{fig:cubic_scaling}
\end{figure}

\section{Scalability}
While the calculation of our loss function scales cubicly in time and space, GPU implementations of matrix multiplication are highly optimized. Consequently, our contextual loss remains practical even for batch sizes as high as 6400. See Figure \ref{fig:cubic_scaling}. In comparison, AP surrogates such as Roadmap become impractical beyond batch sizes of about 2000. We used one A40 GPU for this experiment, so we were unable to plot points above 32 GB of GPU memory.

\begin{figure}
\centering
\includegraphics[width=1.\linewidth]{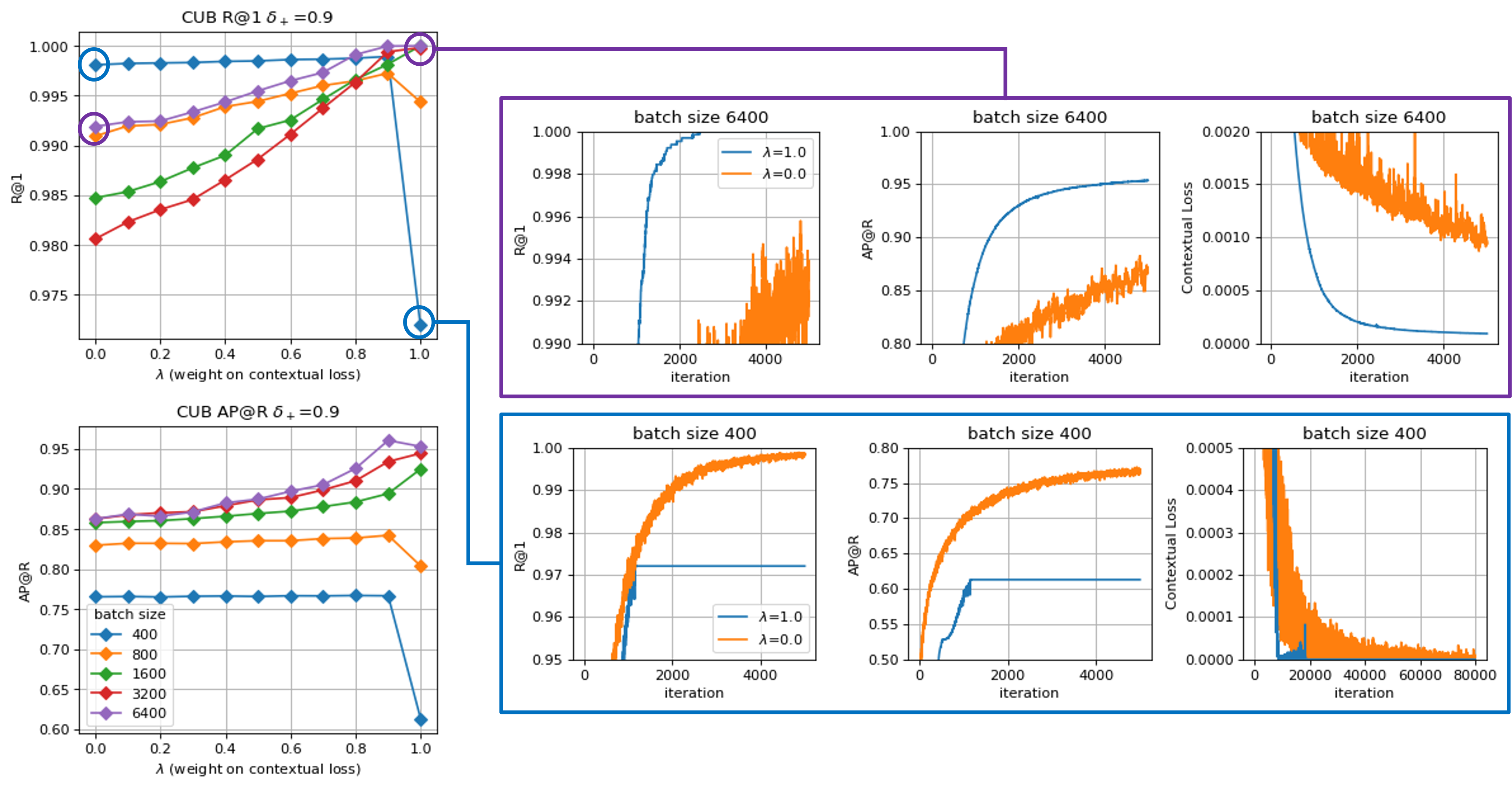}
\caption{Empirical verification of the existence of a decomposability gap when optimizing the contextual loss. We run this experiment on CUB. We train the linear embedding layer on top of fixed ResNet-50 features. We experiment with different $\lambda$. $\lambda=0$ corresponds to the contrastive loss, while $\lambda=1$ corresponds to the contextual loss. We experiment with different batch sizes. We only plot the train R@1 accuracy in this figure. Note that batch size of 6400 corresponds to the entire dataset (i.e. decomposability gap is zero). The plots on the right show the R@1, AP@R, and contextual loss over the course of training for $\lambda=0$ and $\lambda=1$ for both the largest batch size and the smallest batch size. Observe that when the batch size is 400, optimizing the contextual loss ($\lambda=1$) in blue does not converge to the correct ranking on the entire dataset. The training stalls at around 1000 iterations. This is because the embedding space after iteration 1000 is good enough in the sense that the batch-wise ranking is perfect for all sampled batches. Meanwhile, if we use batch size 6400 (no mini-batching), the decomposability gap is zero, and the contextual loss in blue clearly converges to the correct ranking over the entire dataset. Further observe that on the larger batch sizes, the optimal train R@1 is achieved when $\lambda=1$. This suggests that the contrastive loss is not needed on large batch sizes, where the decomposability gap is smaller.}
\label{fig:decomp_gap}
\end{figure}

\section{Decomposability Gap}
\citet{ramzi2021robust} demonstrated mathematically the idea of a decomposability gap in their Roadmap paper. When optimizing a ranking metric such as AP, the average batch-wise AP is a loose upper bound of AP over the entire dataset. Given a reasonable embedding space (e.g. halfway through training), most randomly sampled batches are perfectly ranked, even if ranking over the entire dataset is far from optimal. \citet{ramzi2021robust} call this difference between the batch-wise objective and the dataset-wise objective the ``decomposability gap'' and propose to use the standard contrastive loss to reduce this gap. The resulting optimization objective is a convex combination of the ranking loss and $\L_{\text{contrast}}$. 

This decomposability gap could partially explain why we need to regularize $\L_{\text{context}}$ with $\L_{\text{contrast}}$. Specifically, $\L_{\text{context}}$ reaches a minimum of zero when all samples within a mini-batch are correctly ranked. At this point, the embedding no longer changes until an incorrectly-ranked mini-batch is sampled. This is problematic, because, without hard off-line sampling, the probability that a randomly sampled batch contains the ``hard'' triplet that is incorrectly ranked diminishes. This justifies the need for $\L_{\text{contrast}}$ to complement our contextual loss. See Figure \ref{fig:decomp_gap} for an empirical verification of the existence of a decomposability gap on the CUB dataset.  

\section{Contextual Similarity Definition in Set Notation}
In this section, we state the same contextual similarity definition as Section 3 in the main paper, but following the notation of \citet{kim2022self}. 

We compute the contextual similarity on a single batch, not the entire dataset. Two samples are contextually similar if they share the same neighbors, i.e. the intersection of their $k$-neighbor sets is large. Following this intuition, we calculate $\tilde{w}_{ij}$, the preliminary contextual similarity between samples $i$ and $j$: 
\begin{equation}
    \N_{k+\epsilon}(i) = \{ j \: | \: s_{ij} \le s_{ip} + \epsilon \text{ where $p$ denotes the $k$-th closest neighbor of $i$}\}
\label{eq:appendix_contextsim1}
\end{equation}
\begin{equation}
    \tilde{w}_{ij} = \begin{cases} 
      |\N_{k+\epsilon}(i) \cap \N_{k+\epsilon}(j) | \: / \: |\N_{k+\epsilon}(i)| & , \: \text{if } j \in \N_{k+\epsilon}(i) \\
      0 & , \text{ otherwise.} 
   \end{cases}
\label{eq:appendix_contextsim2}
\end{equation}
We include $i$ in $\N_{k+\epsilon}(i)$ (so if $k=2$, then $\N_{k+0}(i)$ includes two elements: $i$ and its closest neighbor).
Then, we use query expansion and symmeterize to obtain the final contextual similarity. Query expansion \cite{arandjelovic2012three} is a standard evaluation-time trick in metric learning. It boosts retrieval performance by retrieving neighbors of a sample's neighbors. Analogously, we adjust the contextual similarity by averaging $\tilde{w}_{ij}$ over the set of close reciprocal neighbors $\R_{k/2}(i)$. 
\begin{equation}
    \N_{k/2 + \epsilon}(i) = \{ j \: | \: s_{ij} \le s_{ip} + \epsilon \text{ where $p$ denotes the $k/2$-th closest neighbor of $i$}\}
\label{eq:appendix_contextsim3}
\end{equation}
\begin{equation}
    \R_{k/2 + \epsilon}(i) = \{ j \: | \: j \in \N_{k/2 + \epsilon}(i) \text{ and } i \in \N_{k/2 + \epsilon}(j) \}
\label{eq:appendix_contextsim4}
\end{equation}
\begin{equation}
\hat{w}_{ij} = \frac{1}{|\R_{k/2 + \epsilon}(i)|} \sum_{p \in \R_{k/2 + \epsilon}(i)} \tilde{w}_{pj} \:\:\: ,\:\:\: w_{ij} = \frac{1}{2} (\hat{w}_{ij} + \hat{w}_{ji}).
\label{eq:appendix_contextsim5}
\end{equation}
Note that $w_{ij} \in [0,1]$ and only depend on the cosine similarities $s_{ij}$ between embeddings. We want to optimize the embeddings such that $w_{ij}$ converges to $y_{ij}$. The value of $k$ is not arbitrary; it must be set to the number of samples per label in the mini-batch. For example, a standard metric learning setup is to randomly sample 32 labels and then sample 4 images per label. In this case, we set $k=4$.

\subsection{Differences with STML}

This subsection enumerates in detail the technical differences between our definition of contextual similarity ($w_{ij}$ in Eq. \ref{eq:appendix_contextsim5}) and the definition of contextual similarity in STML \citep{kim2022self}. Note that by necessity, we reference equations and notation in the STML paper.
\begin{enumerate}
    \item STML averages a non-linear function of the cosine similarity ($w^P_{ij}$ in their Eq. 5) with the contextual similarity ($w^C_{ij}$ in their Eq. 5) to arrive at an estimate of the ground-truth similarity. In their work, the resulting similarity is used to ``teach'' a student embedding network. In our work, the contextual similarity is optimized directly by the contextual loss $\mathcal{L}_{\text{context}}$, and the cosine similarity is directly optimized by the contrastive loss $\mathcal{L}_{\text{contrast}}$.
    \item We add a margin term $\epsilon$ to the neighborhood definition; STML uses $\epsilon=0$. In our loss, the margin is necessary to encourage separation between embeddings with different labels.
    \item (Minor) We use the set of $k+\epsilon$ neighbors to calculate $\tilde{w}_{ij}$ in our Eq. \ref{eq:appendix_contextsim2}; STML uses the set of $k$ \emph{reciprocal} neighbors to calculate $\tilde{w}_{ij}$ in their Eq. 7.
    \item (Minor) We use the set of $k$ \emph{reciprocal} neighbors for query expansion in our Eq. \ref{eq:appendix_contextsim4}; STML uses the $k/2$ neighbors in their Eq. 8.
\end{enumerate}

Additionally, our work contains the following technical innovations not found in \citep{kim2022self}:
\begin{itemize}
    \item The implementation of contextual similarity in \citep{kim2022self} involves indexing a matrix and setting those values to a constant. Consequently, the output depends on the input \emph{indirectly} through an indexing operation. We re-implement the contextual similarity calculation in a way that only uses greater-than, logical-and, addition, and multiplication. A detailed explanation of this process is detailed in Appendix E.2 and E.3.
    \item We determine suitable differentiable substitutions for the logical-and and greater-than functions. logical-and could be replaced by simple averaging, multiplication, or $\min$. greater-than can be replaced by a sigmoid, a smooth upper-bound, or the constant-gradient heaviside. We ultimately chose to use multiplication and a constant-gradient heaviside. These two choices are intuitively motivated in Appendix E.2, Fig. \ref{fig:figure2}, and experimentally justified in Table \ref{tab:ablation-2}.
    \item (Minor) Optimizing the intersection of neighborhood sets in Eq. \ref{eq:appendix_contextsim2} directly is problematic and leads to shrinkage of the embedding space. We mitigate this issue by additionally optimizing the intersection of the complements of the neighborhood sets. This is explained in Figure \ref{fig:distribution_plots}
\end{itemize}
 
\subsection{Detailed Optimization Motivation}

The definition in the previous sub-section clearly contains three discrete operations: (1) greater-than, (2) logical-and, and (3) intersection. For optimization, we will deal extensively with indicator matrices: we denote as $\ind_\N \in \RR^{n \times n}$ the indicator matrix where $\ind_\N (i,j) = 1 \: \text{if} \: j \in \N(i)$ for some set $\N$. We use $\odot$ to denote element-wise multiplication.


\noindent \textbf{Greater-than} This is used in Eq. \ref{eq:appendix_contextsim1} and \ref{eq:appendix_contextsim3} and is equivalent to the non-differentiable heaviside function $\theta$. Our approach is to use the exact value of $\theta(\cdot)$ in the forward pass and a constant positive gradient $\alpha$ in the backward pass. A constant positive gradient is reasonable since $\theta$ is a (non-strictly) increasing function.

\begin{equation}
    \text{Forward:  } \theta(x) = 1 \text{ if } x \ge 0 \: ; \: 0 \text{ otherwise } \:\:\:\: \text{Backward:  } \frac{\partial \theta(x)}{\partial x} = \alpha.
\label{eq:appendix_theta_hacked}
\end{equation}
In Section F.1 we show with a toy experiment that this approach is robust despite being somewhat heuristic. In contrast, $\theta$ is traditionally approximated by a sigmoid (e.g. for AP approximation and in Gated Recurrent Networks \cite{cho2014properties} ):

\begin{equation}
\theta_{\sigma}(x) = \frac{1}{1 + \exp \left(x / \tau \right)}.
\label{eq:appendix_theta_sgimoid}
\end{equation}
$\theta_{\sigma}$ trades off the quality of the approximation with the domain where gradients are non-zero. As temperature $\tau$ decreases, $\theta_{\sigma}$ approaches $\theta$, but gradient vanishes everywhere except in a small region around the boundary. This behavior is not intuitive: it is undesirable to only have a large gradient at the boundary. Some prior work (e.g. ROADMAP) side-step this issue by using an upper-bound to the heaviside function (where the right side of the heaviside function increases linearly), which solves the gradient issue at the expense of grossly over-estimating the true objective. In our case, this is especially concerning since we will be multiplying together indicator functions. In Section G.1, we show that our approach in Eq. \ref{eq:appendix_theta_hacked} achieves better empirical results than a sigmoid approximation.

\noindent \textbf{Logical-and } A logical-and is used explicitly in Eq. \ref{eq:appendix_contextsim4} and implicitly in Eq. \ref{eq:appendix_contextsim2} and \ref{eq:appendix_contextsim5}. There are two differentiable substitutes for logical-and: $\min$ and multiplication. Multiplication is smooth, but has no gradient at the origin. $\min$ is logically consistent on the continuous domain [0,1], but the gradient is not continuous when inputs are equal. We found experimentally that multiplication is the best option, see row 5 of Table \ref{tab:ablation-2}. A gradient of zero at the origin is desireable in the case of Eq. \ref{eq:appendix_contextsim4} and \ref{eq:appendix_contextsim5} (query expansion step). A work-around for the zero-gradient issue for Eq. \ref{eq:appendix_contextsim2} is presented in the next sub-section.

\noindent \textbf{Intersection } The number of elements in the intersection in Eq. \ref{eq:appendix_contextsim2} is calculated using matrix multiplication. A sample $p \in \N_{k+\epsilon}(i) \cap \N_{k+\epsilon}(j)$ if $p \in \N_{k+\epsilon}(i) \text{ and } p \in \N_{k+\epsilon}(j)$. Using multiplication for the logical-and, the number of elements in the intersection can be expressed compactly as a function of the indicator matrices:
\begin{equation}
    M_+ = \ind_{\N_{k+\epsilon}} \ind_{\N_{k+\epsilon}}^\intercal \: , \: \text{where} \:  M_+(i,j) = \left|\N_{k+\epsilon}(i) \cap \N_{k+\epsilon}(j)\right|.
\end{equation}
We are now ready to state the loss function by combining the definition of contextual similarity with the optimization method.

\begin{figure}
\centering
\includegraphics[width=1.\linewidth]{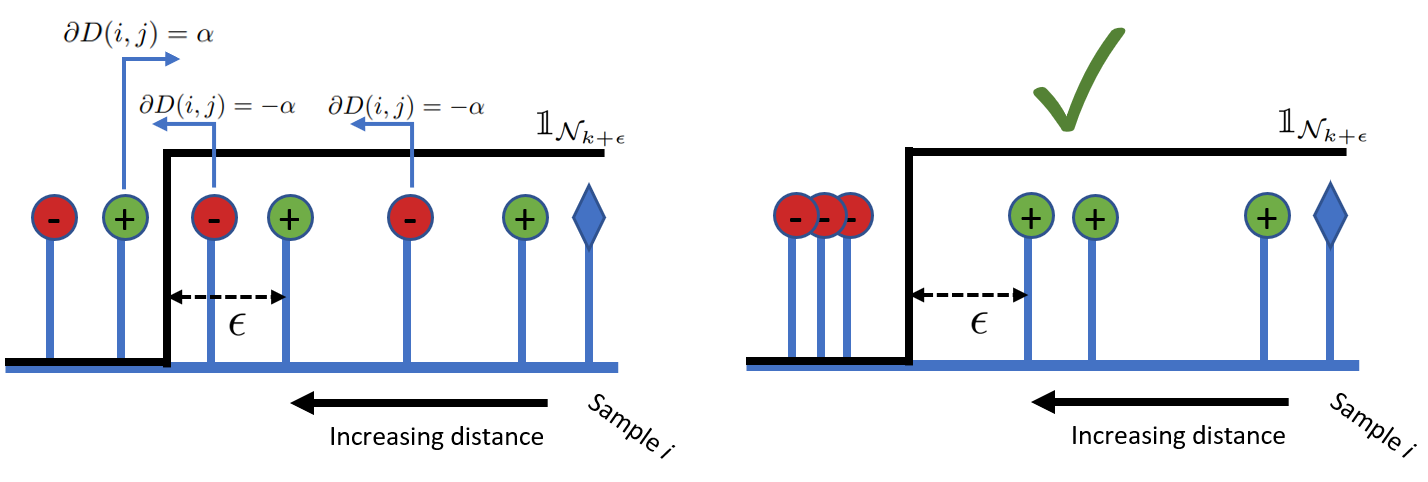}
\caption{Illustration of $\N_{k+\epsilon}$ optimization using the simplified loss function $\L_1$ in Eq. \ref{eq:appendix_ell_1}. Negative samples which are in the set $\N_{k+\epsilon} (i)$ are pushed away from $i$ by a constant gradient. Positive samples which are not in the set $\N_{k+\epsilon} (i)$ are pulled closer. This behavior is analogous to the standard contrastive loss: $\L_{\text{contrast}}$ applies a constant gradient to pairs violating a fixed margin, while $\L_1$ applies a constant gradient to pairs violating the flexible margin defined by the $k$-th neighbor. $k=4$ in the illustration.}
\label{fig:figure2}
\end{figure}

\subsection{Loss in Matrix Notation}

Our loss function straightforwardly combines the previous two sub-sections, with one exception for the intersection calculation in Eq. \ref{eq:appendix_contextsim2}. Optimizing the intersection with $M_+$ tends to focus on pairs of neighboring samples because using multiplication for logical-and zeros out the gradient when both inputs are 0. This is problematic since the resulting embedding becomes clustered regardless of true similarity (see Figure \ref{fig:distribution_plots}). 
We mitigate this problem by optimizing the intersection of the complements $\N_{k+\epsilon}^c(i) \cap \N_{k+\epsilon}^c(j)$ in addition to optimizing the original intersection in Eq. \ref{eq:appendix_contextsim2}. $^c$ denotes the complement of a set. Maximizing the size of the intersection between two sets is equivalent to maximizing the size of the intersection between their complements, and vice versa. This can be trivially proven under the assumptions that the universal set is constant and that the size of both sets is constant.

\begin{figure}
\centering
\includegraphics[width=1.\linewidth]{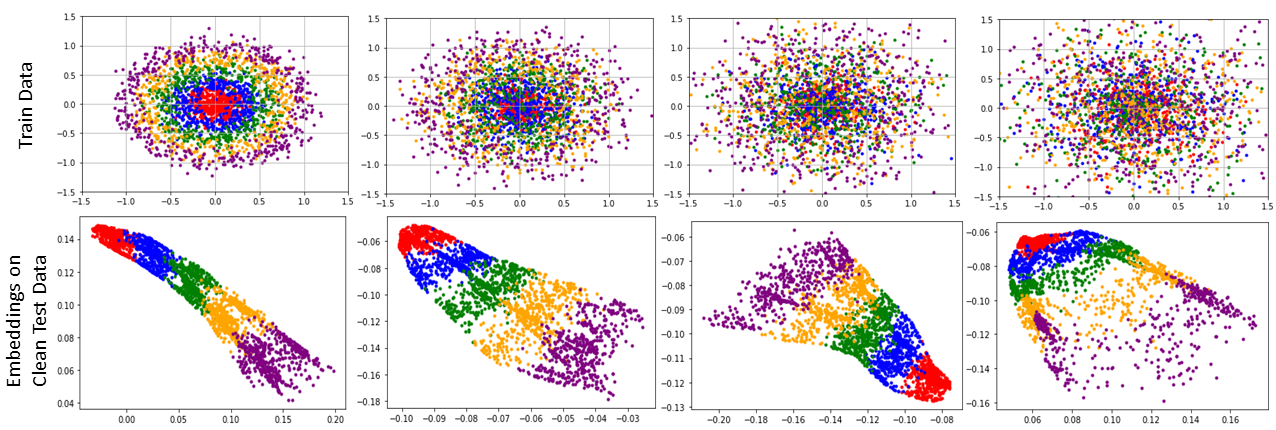}
\caption{This figure justifies the non-standard approach of optimizing $\L_1$ (Eq. \ref{eq:appendix_ell_1}) by overriding the gradient of $\theta$. We generate synthetic 2-D data consisting of five concentric circles; each color represents a label. We use a standard three layer MLP with ReLU activations to output a 2D embedding. We train on increasingly noisy data and plot the resulting embeddings on clean data. We only sample 4 points per label per batch ($k=4$). $\L_1$ minimization results in reasonable embeddings despite its simplicity. }
\label{fig:toy_experiment}
\end{figure}

Let $D \in \R^{n \times n}$ denote the matrix where $D(i,j) \in [0,4]$ is the squared Euclidean distance between the normalized features of sample $i$ and $j$. $D(i,j) = 2 - 2 s_{ij}$. $\sg$ denotes stop gradient.

\noindent \textbf{Step 1 (Neighborhood Optimization):}
\begin{equation}
    \ind_{\N_{k+\epsilon}}(i,j) = \theta(-D(i,j) + \sg(D(i, p)) + \epsilon) \text{ where $p$ denotes the $k$-th closest neighbor of $i$.}
    \label{eq:appendix_loss1}
\end{equation}
\noindent \textbf{Step 2 (Optimizing Intersection of Neighborhoods):}
\begin{equation}
\begin{split}
        \tilde{W} = \frac{1}{2} \left(\frac{M_+}{\sg(|\N_{k+\epsilon}(i)|)} + \frac{M_-}{\sg(|\N_{k+\epsilon}^c(i)|)} \right) \odot \ind_{\N_{k+\epsilon}} \: \\ \text{where} \:\: M_+ = \ind_{\N_{k+\epsilon}} \ind_{\N_{k+\epsilon}}^\intercal \text{  and  } M_- = \ind_{\N_{k+\epsilon}^c} \ind_{\N_{k+\epsilon}^c}^\intercal.
\end{split}
\label{eq:appendix_loss2}
\end{equation}

\noindent \textbf{Step 3 (Query Expansion):}
\begin{equation}
    \ind_{\N_{k/2 + \epsilon}}(i,j) = \theta(-D(i,j) + \sg(D(i, p)) + \epsilon) \text{ where $p$ denotes the $k/2$-th closest neighbor of $i$}
    \label{eq:appendix_loss3}
\end{equation}
\begin{equation}
    \ind_{\R_{k/2 + \epsilon}} = \ind_{\N_{k/2 + \epsilon}} \odot \ind_{\N_{k/2 + \epsilon}}^\intercal
    \label{eq:appendix_loss4}
\end{equation}
\begin{equation}
    \hat{W} = \frac{\ind_{\R_{k/2 + \epsilon}} \tilde{W} }{ |\R_{k/2 + \epsilon}(i)| } \: , \: W = \frac{1}{2} \left( \hat{W} + \hat{W}^\intercal \right).
    \label{eq:appendix_loss5}
\end{equation}
Finally, we use the MSE loss to optimize $w_{ij} := W(i,j)$ against the true similarity labels $y_{ij}$:
\begin{equation}
    \L_{\text{context}} = \frac{1}{n^2} \sum_{i,j|i \ne j} ( y_{ij} - w_{ij} )^2.
    \label{eq:appendix_loss7}
\end{equation}
$\L_{\text{context}}$ is the key to our framework and works reasonably on its own. However, it has two vulnerabilities: (1) similar to learning to rank losses, $\L_{\text{context}}$ can be small even if the distance between positive pairs is large, so long as there are no closer negative pairs (see Figure \ref{fig:figure2}); (2) $\L_{\text{context}}$ tends to converge to a solution where the average cosine similarity between all pairs is large, suggesting that only a small portion of the available embedding space is utilized. In response to problem (1), we add the standard contrastive loss (\cite{hadsell2006dimensionality}), which explicitly optimizes the cosine similarity toward fixed margins $\delta_{+}$ and $\delta_{-}$. In response to problem (2), we add a non-standard but straightforward similarity regularizer, which regularizes the average cosine similarity between all pairs toward a fixed value $\tilde{s}$. Our final framework (Eq. \ref{eq:appendix_loss6}) minimizes a weighted combination of the three losses.
\begin{equation}
    \L_{\text{contrast}} = \frac{\sum_{i,j | y_{ij} = 1} (\delta_{+} - s_{ij} )_{+} }{\big|\{i,j | y_{ij} = 1 \text{ and } \delta_{+} - s_{ij} > 0\}\big|}  + \frac{\sum_{i, j | y_{ij} = 0} (s_{ij} - \delta_{-})_{+}}{\big|\{i,j | y_{ij} = 0 \text{ and } s_{ij} - \delta_{-} > 0\}\big|} 
    \label{eq:appendix_loss8}
\end{equation}
\begin{equation}
      \L_{\text{reg}} =  \left( \tilde{s} - \frac{1}{n^2} \sum_{i,j}^{n^2} s_{ij} \right)^2
    \label{eq:appendix_regularizer}
\end{equation}
\begin{equation}
    \L_{\text{ours}} = \lambda \L_{\text{context}} + (1 - \lambda) \L_{\text{contrast}} + \gamma  \L_{\text{reg}}.
    \label{eq:appendix_loss6}
\end{equation}

Note that \citet{el2021training} propose a superficially similar similarity regularization scheme, which ``maximizes the distance between every point and its nearest neighbor''. Our $\L_{\text{reg}}$ regularizes the average similarity between all pairs of samples. Nearest neighbor pairs are dominated by positive pairs, while all pairs are dominated by negative pairs. Therefore, our regularizer is fundamentally different from \citet{el2021training}.


\section{Slightly Different Analysis}

The contextual similarity loss function $\L_{\text{context}}$ is highly non-trivial. In this section we demystify each of the three steps using a toy experiment and gradient analysis. This is a slightly different analysis than the one given in the main paper.

\subsection{Step 1: Neighborhood optimization}

Consider the following simplified version of $\L_{\text{context}}$ :
\begin{equation}
    \L_1 = \L_{\text{MSE}}(y_{ij}, \ind_{\N_{k+\epsilon}}(i,j)).
    \label{eq:appendix_ell_1}
\end{equation}
This is a valid loss function that works reasonably on its own (see Figure \ref{fig:toy_experiment}). During each iteration, we sample a batch with exactly $k$ samples from each label. Intuitively, $\L_1$ reaches a minimum of zero when $\N_{k+\epsilon}(i)$ includes only $i$ and the $k-1$ other samples with the same label, i.e. all samples are correctly ranked. Using the gradient of $\theta(\cdot)$ defined in Eq. \ref{eq:appendix_theta_hacked}, we see that negative samples which are in $\N_{k+\epsilon}(i)$ receive a gradient of magnitude $\alpha$ pushing it away from $i$, while positive samples which are not in $\N_{k+\epsilon}(i)$ receive a gradient of magnitude $\alpha$ pushing it toward $i$. See Figure \ref{fig:figure2}.

\begin{figure}
\centering
\begin{subfigure}{.48\textwidth}
  \centering
  \includegraphics[width=1.\linewidth]{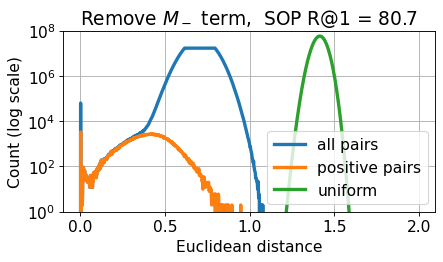}
\end{subfigure}
\begin{subfigure}{.48\textwidth}
  \centering
  \includegraphics[width=1.\linewidth]{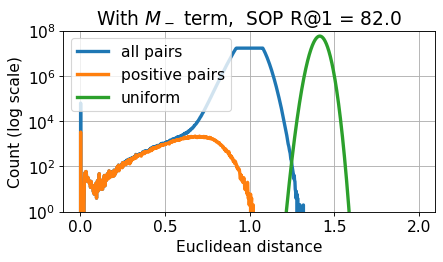}
\end{subfigure}
\caption{This figure justifies the $M_-$ term in Eq. \ref{eq:appendix_loss2}. The left plot shows the distribution of distances between pairs in normalized embedding space when we only optimize $M_+$; the right plot shows the same when we optimize both $M_-$ and $M_+$. A uniformly distributed embedding space has an average pair-wise distance of $\sqrt{2}$ because most directions are orthogonal in high dimensions. This is indicated by the green distribution. Clearly, more of the embedding space is utilized when we optimize both $M_-$ and $M_+$. }
\label{fig:distribution_plots}
\end{figure}

\subsection{Step 2: Optimizing Intersection of neighborhoods}

While $ \lambda \L_1 + (1 - \lambda) \L_{\text{contrastive}}$ is a reasonable loss function already, row 1 of Table \ref{tab:ablation-2} shows that $\L_1$ collapses without the contrastive loss ($\lambda = 1$). We propose that this is because $\L_1$ provides very sparse gradients: $\partial \L_1 / \partial s_{ij} = 0$ for most $(i,j)$ pairs. Consider the following more complicated loss:
\begin{equation}
    \L_2 = \L_{\text{MSE}}(y_{ij}, \tilde{W}(i,j)).
    \label{eq:appendix_ell_2}
\end{equation}
$\tilde{W}$ is defined in Eq. \ref{eq:appendix_loss2}. $\tilde{W}(i,j)$ has two terms multiplied together element-wise. The $\ind_{\N_{k+\epsilon}}$ term was already addressed in the previous sub-section, so we focus on the first term, which optimizes the intersection between neighborhood sets $M_+$. We offer a straight-forward intuition: \emph{Maximizing the intersection between the neighborhood sets of two samples is equivalent to pushing one sample towards the context of the other sample and vice versa}; \emph{minimizing the intersection is equivalent to pulling apart the contexts of the two samples}. We show this intuition by analyzing the gradient.

For simplicity, consider $\epsilon = 0$, such that the normalization factors in Eq. \ref{eq:appendix_loss2} are equal to $k$ and $n-k$, resp. Further consider a negative pair of samples $i$ and $j$  ($y_{ij} = 0$), where $j$ is wrongly ranked w.r.t. $i$ (i.e. $j \in \N_{k}(i)$). Following the loss function equations:
\begin{equation}
\begin{split}
     M_+ (i,j) &= \langle \ind_{\N_k} (i), \ind_{\N_k} (j) \rangle \:\:\:\: \:\:\:\:
     M_- (i,j) =  \langle \ind_{\N_k}^c (i), \ind_{\N_k}^c (j) \rangle
     \\
     \tilde{W} (i,j) &= \frac{1}{2} \left( \frac{1}{k} M_+ (i,j) + \frac{1}{n-k} M_- (i,j) \right) \cdot \underbrace{\ind_{\N_k}(i,j)}_{=1}.
\end{split}
\end{equation}
$\tilde{W} (i,j) \in (0,1] $ must be a non-zero positive number. Using the equation for $\L_2$, we see that the gradient w.r.t. $\tilde{W} (i,j)$ must be positive because $y_{ij} = 0$. We are now ready for the backward pass. For simplicity of notation, $\partial \tilde{W} (i,j) := g_{ij} > 0$ denotes the gradient of the loss w.r.t. $\tilde{W} (i,j)$.
\begin{equation}
    \begin{split}
        \partial  M_+ (i,j) =  \frac{1}{2k} g_{ij}, &\text{ and } \partial  M_- (i,j) = \frac{1}{2(n-k)} g_{ij}
        \\
        \partial \ind_{\N_k} (i) = \frac{1}{2k}  g_{ij} \ind_{\N_k} (j) \: &, \:  \partial \ind_{\N_k} (j) = \frac{1}{2k}  g_{ij} \ind_{\N_k} (i)
        \\
        \partial \ind_{\N_k}^c (i) = \frac{1}{2(n-k)}  g_{ij} \ind_{\N_k}^c (j) \: &, \:  \partial \ind_{\N_k}^c (j) = \frac{1}{2(n-k)} g_{ij} \ind_{\N_k}^c (i).
    \end{split}
\end{equation}
Use the fact that $\ind_{\N_k^c} = 1 - \ind_{\N_k}$ and $\partial \ind_{\N_k} = -\partial \ind_{\N_k^c}$ and vice versa:
\begin{equation}
    \begin{cases}
      \partial \ind_{\N_k} (i) &= \frac{g_{ij}}{2k} \left( \ind_{\N_k} (j) \right) \\
      \partial \ind_{\N_k} (j) &= \frac{g_{ij}}{2k} \left( \ind_{\N_k} (i)  \right) \\
    \end{cases}
\label{eq:appendix_foo1}
\end{equation}
\begin{equation}
    \begin{cases}
      \partial \ind_{\N_k} (i) &= \frac{g_{ij}}{2(n-k)} \left( \ind_{\N_k} (j) -1\right) \\
      \partial \ind_{\N_k} (j) &= \frac{g_{ij}}{2(n-k)} \left( \ind_{\N_k} (i)  -1\right) \\
    \end{cases}
\label{eq:appendix_foo2}
\end{equation}
Eq. \ref{eq:appendix_foo1} is the contribution to the gradient from optimizing $\ind_{\N_k}$, while Eq. \ref{eq:appendix_foo2} is the contribution to the gradient from optimizing $\ind_{\N_k}^c$. These can be added together:
\begin{equation}
    \begin{cases}
      \partial \ind_{\N_k} (i) &= \frac{g_{ij}}{2(n-k)} \left( \ind_{\N_k} (j) -1\right) + \frac{g_{ij}}{2k} \left( \ind_{\N_k} (j) \right)\\
      \partial \ind_{\N_k} (j) &= \frac{g_{ij}}{2(n-k)} \left( \ind_{\N_k} (i)  -1\right) + \frac{g_{ij}}{2k} \left( \ind_{\N_k} (i)  \right)\\
    \end{cases}
\label{eq:appendix_foo3}
\end{equation}
Going backward through $\theta(\cdot)$ multiplies the gradient by $-\alpha$ (negative sign accounts for optimizing distance instead of similarity):
\begin{equation}
    \partial D(i) = \frac{-\alpha g_{ij}}{2(n-k)} \left( \ind_{\N_k} (j) -1\right) - \frac{\alpha g_{ij}}{2k} \left( \ind_{\N_k} (j) \right) \:,\: \text{and vice versa for } \partial D(j).
\end{equation}
From the above equation, $\partial D(i,p) < 0$ when $\ind_{\N_k} (j,p) = 1$ and $\partial D(i,p) > 0$ when $\ind_{\N_k} (j,p) = 0$, for all samples $p$ in the batch. In words, we increase the distance between $i$ and all samples in the neighborhood of $j$, and we decrease the distance between $i$ and all points outside the neighborhood of $j$. Recall that we assumed $i$ and $j$ to be a negative pair, so it makes sense to pull $i$ away from the context of $j$ in this fashion.

\setlength\tabcolsep{6 pt}
\begin{table}
\caption{R@$k$ Results. We use ResNet-50 with an embedding size of 512 for all experiments. $\dagger$ indicates results reported by the original authors; we re-run all other baselines using the implementation by \cite{musgrave2020pytorch}. Standard deviations are based on three trials with the same train-test split. We emphasize that our R@1 performance is at least two standard deviations better than the next best baseline on all datasets. We are at least comparable to baselines on other R@$k$ metrics. } 

\centering
{  \small
\begin{tabular}{ l >{\columncolor[gray]{0.9}}c c c c >{\columncolor[gray]{0.9}} c c c c  }
 \toprule
 & \multicolumn{4}{c}{CUB} & \multicolumn{4}{c}{Cars} \\
 \midrule
 Method & R@1 & R@2 & R@4 & R@8 & R@1 & R@2 & R@4 & R@8  \\
  \midrule
Contrastive & 68.5 $\pm$ 0.3 & 78.3 $\pm$ 0.1 & 86.0 $\pm$ 0.2 & 91.3 $\pm$ 0.1 & 85.4 $\pm$ 0.2 & 91.1 $\pm$ 0.3 & 94.6 $\pm$ 0.3 & 96.8 $\pm$ 0.1 \\
 Triplet & 67.3 $\pm$ 0.2 & 77.9 $\pm$ 0.1 & 85.6 $\pm$ 0.2 & 91.2 $\pm$ 0.1 & 77.6 $\pm$ 1.3 & 85.4 $\pm$ 0.8 & 90.8 $\pm$ 0.7 & 94.1 $\pm$ 0.4 \\
 NtXent & 65.7 $\pm$ 0.4 & 76.3 $\pm$ 0.2 & 84.3 $\pm$ 0.4 & 90.0 $\pm$ 0.4 & 79.0 $\pm$ 0.6 & 86.0 $\pm$ 0.3 & 91.0 $\pm$ 0.2 & 94.4 $\pm$ 0.3 \\
 MS & 68.9 $\pm$ 0.5 & 78.5 $\pm$ 0.4 & 86.0 $\pm$ 0.6 & 91.4 $\pm$ 0.5 & 88.7 $\pm$ 0.4 & 93.0 $\pm$ 0.2 & 95.7 $\pm$ 0.1 & 97.3 $\pm$ 0.1 \\
 N-Softmax$\dagger$ & 61.3 & 73.9 & 83.5 & 90.0 & 84.2 & 90.4 & 94.4 & 96.9 \\
 Proxy NCA ++$\dagger$ & 69.0 $\pm$ 0.8 & \textbf{79.8} $\pm$ 0.7 & \textbf{87.3} $\pm$ 0.7 & \textbf{92.7} $\pm$ 0.4 & 86.5 $\pm$ 0.4 & 92.5 $\pm$ 0.3 & 95.7 $\pm$ 0.2 & \textbf{97.7} $\pm$ 0.1 \\
 Fast-AP & 63.3 $\pm$ 0.1 & 73.7 $\pm$ 0.4 & 82.2 $\pm$ 0.3 & 88.5 $\pm$ 0.2 &  74.7 $\pm$ 0.4 & 82.5 $\pm$ 0.7 & 88.0 $\pm$ 0.6 & 92.2 $\pm$ 0.2 \\
 Smooth-AP & 66.5 $\pm$ 0.9 & 76.6 $\pm$ 0.5 & 84.8 $\pm$ 0.6 & 90.8 $\pm$ 0.4 &  81.1 $\pm$ 0.2 & 87.8 $\pm$ 0.4 & 92.2 $\pm$ 0.3 & 95.1 $\pm$ 0.3 \\
 ROADMAP & 68.7 $\pm$ 0.5 & 78.3 $\pm$ 0.3 & 86.1 $\pm$ 0.3 & 91.1 $\pm$ 0.1 & 84.5 $\pm$ 0.5 & 90.3 $\pm$ 0.0 & 93.9 $\pm$ 0.0 & 96.2 $\pm$ 0.1 \\
 Ours & \textbf{69.8} $\pm$ 0.2 & \textbf{79.8} $\pm$ 0.1 & \textbf{87.1} $\pm$ 0.1 & 92.3 $\pm$ 0.2 & \textbf{89.3} $\pm$ 0.0 & \textbf{93.7} $\pm$ 0.2 & \textbf{96.3} $\pm$ 0.1 & \textbf{97.8} $\pm$ 0.2 \\
 \bottomrule
\end{tabular}
}
\setlength\tabcolsep{8.8 pt}
{  \small
\begin{tabular}{ l >{\columncolor[gray]{0.9}} c c c >{\columncolor[gray]{0.9}} c c c c }
 \toprule
 & \multicolumn{3}{c}{SOP} & \multicolumn{4}{c}{mini-iNaturalist}\\
 \midrule
 Method & R@1 & R@10 & R@100 & R@1 & R@4 & R@16 & R@32\\
  \midrule
Contrastive & 82.4 $\pm$ 0.0 & 91.9 $\pm$ 0.0 & 96.0 $\pm$ 0.0 & 43.5 $\pm$ 0.1 & 62.7 $\pm$ 0.1 & 77.6 $\pm$ 0.1 & 83.2 $\pm$ 0.1 \\
 Triplet & 82.0 $\pm$ 0.0 & 92.5 $\pm$ 0.1 & 96.7 $\pm$ 0.0 & 35.4 $\pm$ 0.1 & 56.5 $\pm$ 0.1 & 74.7 $\pm$ 0.1 & 81.7 $\pm$ 0.1 \\
 NtXent & 79.7 $\pm$ 0.2 & 90.8 $\pm$ 0.0 & 96.1 $\pm$ 0.0 & 40.8 $\pm$ 0.1 & 61.6 $\pm$ 0.1 & 78.0 $\pm$ 0.0 & 83.9 $\pm$ 0.0 \\
 MS & 81.4 $\pm$ 0.0 & 91.4 $\pm$ 0.0 & 96.1 $\pm$ 0.1 & 44.9 $\pm$ 0.1 & 63.9 $\pm$ 0.1 & 78.4 $\pm$ 0.1 & 83.9 $\pm$ 0.1 \\
 N-Softmax$\dagger$ & 78.2 & 90.6 & 96.2 & -- & -- & -- & -- \\
 Proxy NCA ++$\dagger$ & 80.7 $\pm$ 0.5 & 92.0 $\pm$ 0.3 & 96.7 $\pm$ 0.1 & -- & -- & -- & -- \\
 Fast-AP & 80.3 $\pm$ 0.1 & 91.0 $\pm$ 0.1 & 96.0 $\pm$ 0.0 & 35.6 $\pm$ 0.2 & 55.8 $\pm$ 0.1 & 72.8 $\pm$ 0.0 & 79.3 $\pm$ 0.0 \\
 Smooth-AP & 82.0 $\pm$ 0.0 & 92.6 $\pm$ 0.0 & \textbf{96.9} $\pm$ 0.0 & 42.7 $\pm$ 0.0 & 63.3 $\pm$ 0.0 & 79.0 $\pm$ 0.0 & 84.7 $\pm$ 0.0 \\
 ROADMAP & 83.1 $\pm$ 0.1 & 92.6 $\pm$ 0.0 & 96.6 $\pm$ 0.0 & 45.9 $\pm$ 0.1 & \textbf{65.8} $\pm$ 0.0 & \textbf{80.4} $\pm$ 0.1 & \textbf{85.7} $\pm$ 0.0 \\
 Ours & \textbf{83.3} $\pm$ 0.0 & \textbf{92.9} $\pm$ 0.1 & 96.7 $\pm$ 0.0 & \textbf{46.2} $\pm$ 0.0 & \textbf{65.8} $\pm$ 0.1 & 80.2 $\pm$ 0.1 & 85.4 $\pm$ 0.1 \\
 \bottomrule
\end{tabular}
}
\label{tab:main}
\end{table}
\setlength\tabcolsep{6 pt}

\subsection{Step 3: Query Expansion}

Query expansion (QE) is an established trick for image retrieval \cite{arandjelovic2012three}. QE expands the neighborhood set by additionally retrieving the neighbors of very close neighbors. In the unsupervised setting, QE refers to averaging the contextual similarity scores for samples in the $\R_{k/2 + \epsilon}$ neighborhood (see Eq. \ref{eq:appendix_loss5}); this leads to more robust pseudo-supervision. In our setting, we have two reasons for using QE (Eq. \ref{eq:appendix_loss3} \ref{eq:appendix_loss4} and \ref{eq:appendix_loss5}): (1) averaging $\tilde{w}_{ij}$ with very close neighbors could have the same effect as label smoothing and (2) the $\N_{k/2 + \epsilon}$ neighborhood is optimized by Eq. \ref{eq:appendix_loss4} and \ref{eq:appendix_loss5}. Empirically, QE is necessary to achieve the optimal performance of our framework, since $\L_{\text{context}}$ achieves higher R@1 than $\L_2$ in Table \ref{tab:ablation-2}.

\section{256 $\times$ 256 Resolution Experiments}
We include 256 $\times$ 256 image resolution results in this section. This setup is slightly different from the main paper. These results are comparable to the results reported in the ROADMAP paper.

\noindent \textbf{Hyperparameters and Setup on 256 $\times$ 256 Experiments } We use hyperparameter values that are approximately optimal across all datasets. $\lambda=0.4$, $\gamma=0.1$, $\alpha=10.0$, $\epsilon=0.05$, $k=4$, $\delta_+ = 0.75$, $\delta_- = 0.6$, $\tilde{s} = 0.25$. We follow the same training procedure as ROADMAP, but with a slightly faster learning-rate schedule for time efficiency. We use Adam with a learning-rate schedule that multiplies the learning rate by 0.3 at Epochs 15, 30, and 45. We train for 80 epochs. We report results on the model with the best test R@1 metric, as is standard in the literature. We use an initial learning rate of $8 \times 10^{-5}$ on CUB, $0.00016$ on Cars, $4 \times 10^{-5}$ on SOP, and $8 \times 10^{-5}$ on iNaturalist. We use a batch size of 256 for iNaturalist and 128 for other datasets; the larger batch size is necessary to achieve reasonable performance on iNaturalist. We use a 4 per class sampler. For CUB and Cars, we use random sampling (sample 32 classes at random, then sample 4 images per class). For SOP and iNaturalist, we use hierarchical sampling (\citet{cakir2019deep}), following prior work. We use an embedding size of 512 and ResNet-50 with a linear embedding layer. Following prior work, we use layer-norm and max-pooling on the smaller CUB and Cars datasets. We always freeze batch-norm.

\noindent \textbf{Discussion for 256 $\times$ 256 Experiments } Our R@1 results are at least two standard deviations better than the best baseline across all datasets. Our results are especially good on CUB and Cars, where we achieve R@1 gains of 0.8\% and 0.6 \%, resp. We achieve more modest R@1 gains of 0.2\% and 0.3\% on SOP and iNaturalist, resp. We note that these gains are significant because the standard deviation is relatively small. We also note that while Proxy NCA ++ and MS are the best baselines on CUB and Cars, ROADMAP is the best baseline on SOP and iNaturalist. Our method is the best across \emph{all} datasets, suggesting that it is more versatile.

\begin{table}
\caption{Ablation Results. Here, we experiment with variations of $\L_{\text{context}}$, both by itself ($\lambda=1$) and regularized by a small amount of contrastive loss ($\lambda=0.8$). We exhaustively test various ways to simplify or modify the contextual loss presented in Eq. \ref{eq:appendix_loss1} - \ref{eq:appendix_loss7}. On the SOP dataset, we show that all of the modifications to $\L_{\text{context}}$ decrease R@1. }
\centering
{ 
\begin{tabular}{ l l c c r }
 \toprule
 && \multicolumn{2}{c}{SOP R@1} \\
 \multicolumn{2}{l}{Ablation ($\gamma=0$)} & $\lambda=0.8$ & $\lambda=1.0$ & Explanation \\
 \midrule
$\lambda \L_{1}$ & $+ (1 - \lambda) \L_{\text{contrast}}$ & 83.13 & 41.99 & Eq. \ref{eq:appendix_ell_1} (step 1 only) \\
$\lambda \L_{1,\sigma}$ & $+ (1 - \lambda) \L_{\text{contrast}}$ & 79.20 & 75.61 & Eq. \ref{eq:appendix_ell_1} but using $\theta_\sigma$ in Eq. \ref{eq:appendix_theta_sgimoid} \\
$\lambda \L_{2}$ & $ + (1 - \lambda) \L_{\text{contrast}}$ & 82.39 & 81.05 & Eq. \ref{eq:appendix_ell_2} (skip step 3) \\
\multicolumn{2}{l}{$\tilde{W} = \ind_{\N_{k + \epsilon}} $  (skip step 2)} & 82.74 & 81.74 & \\
$\lambda \L_{\text{context}, \min}$ & $ + (1 - \lambda) \L_{\text{contrast}}$ & 66.57 & -- & Use $\min$ instead of $\odot$ for logical-and\\
$\lambda \L_{\text{context}, \sigma} $ & $+ (1 - \lambda) \L_{\text{contrast}}$ & 82.39 & 77.84 & Use $\theta_\sigma$ for all steps \\
$\lambda \L_{\text{context}, M_+}  $ & $+ (1 - \lambda) \L_{\text{contrast}}$ & 82.31 & 80.72 & Remove $M_-$ term from Eq. \ref{eq:appendix_loss2} \\
\multicolumn{2}{l}{No stop gradient in Eq. \ref{eq:appendix_loss2}} & 73.03 & -- & \\
$\lambda \L_{\text{context}}$ & $+ (1 - \lambda) \L_{\text{contrast}}$ & \textbf{83.20} & \textbf{82.04} & Final results without $\L_{\text{reg}}$ \\
 \bottomrule
\end{tabular}
}
\label{tab:ablation-2}
\end{table}

\subsection{Ablation Results}
In Table \ref{tab:ablation-2}, we evaluate the contribution of each step in the calculation of $\L_{\text{context}}$. These experiments focus on dissecting $\L_{\text{context}}$, so we set $\gamma=0$ (no similarity regularizer) and $\lambda=0.8$ or $1.0$. Overall, all of the modifications tested in Table \ref{tab:ablation-2} decrease the performance of $\L_{\text{context}}$ in terms of R@1. In particular: row 1 shows that including only step 1 leads to a collapsed representation when $\lambda=1.0$; rows 3 and 4 show that steps 3 and 2 of the loss calculation are necessary; row 7 shows that the $M_-$ term in Eq. \ref{eq:appendix_loss2} is necessary; rows 2 and 6 show that our approach to optimizing $\theta$ in Eq. $\ref{eq:appendix_theta_hacked}$ is better than using a sigmoid.

\section{Minor Experimental Details}

\subsection{Code and Environment}
We include code with our submission. We run experiments on 1 V100 GPU with 16 GB of memory. The CUB and Cars experiments take under one hour. The SOP and iNaturalist experiments take 4 hours and 6 hours, respectively. Some of our code is borrowed from ROADMAP. For faster experimentation, we use mixed precision floating point. Experiments take more than 12 hours on P100 GPUs, partially because mixed precision arithmetic does not appear to speed-up experiments as much on P100 GPUs compared to on V100 GPUs.

\subsection{Augmentation (Specific to 256 $\times$ 256 Experiments)} On CUB, Cars and SOP, we use random resized crop with default parameters to crop the image to 256 $\times$ 256 pixels. We horizontally flip the image with 50\% probability. For testing, we resize the image to 288 pixels, then center crop to 256 pixels. On iNaturalist, we random resize crop to 224 $\times$ 224 pixels, then flip horizontally with 50\% probability. We use a smaller image size on iNaturalist so that the larger batch size can fit in the 16 GB of GPU memory. For testing, we resize the image to 256 pixels then center crop to 224 pixels.

\subsection{Sampling (Specific to 256 $\times$ 256 Experiments)} We use a batch size of 128 on CUB, Cars and SOP. We use a batch size of 256 on iNaturalist. On all datasets, we train for 80 epochs. An epoch is defined as 15, 30, 655, and 576 batches for CUB, Cars, SOP, and iNaturalist respectively. On CUB and Cars, we use a random 4 per class sampler. On SOP and iNaturalist, we use a balanced hierarchical sampling strategy, since there are super-labels. In particular, half of each batch is randomly sampled from one super-label, and the other half is randomly sampled from another super-label. In order to sample in a balanced manner (to prevent over-emphasis of uncommon super-labels), we arrange all samples in the training dataset into half-batches with same super-labels at the beginning of each epoch. We then form batches by pairing up half-batches in a round-robin fashion.

\subsection{Loss Plot } Figure \ref{fig:loss_plot} plots the contextual loss and test R@1 over the course of training on SOP, with varying $\lambda$. $\gamma=0$. These plots show that the loss decreases over the course of training; in general, a lower contextual loss corresponds to a higher test R@1. This result suggests that the contextual loss is a reasonable objective for image retrieval.

\subsection{Note on Contrastive Loss } The contrastive loss has two margin parameters $\delta_-$ and $\delta_+$. The optimal values for $\delta_+$, the positive margin, is different depending on the value of $\gamma$ (how much similarity regularization is used). For the ``contrastive'' results in row 1 of Table \ref{tab:main}, we use $\delta_- = 0.6$ and $\delta_+ = 0.9$. For our results where $\gamma = 0.1$, we set $\delta_- = 0.6$ and $\delta_+ = 0.75$. We find this tighter positive margin to be optimal under similarity regularization. However, we note that $\delta_+ = 0.75$ is likely to be too small when only the contrastive loss is used, since positive pairs are not encouraged to be more similar than a cosine similarity of 0.75.

\subsection{Note on Stop Gradient } Note that we detach the normalization factors in Eq. \ref{eq:appendix_loss2}. This is necessary because optimizing the size of the neighborhood is undesirable. On the contrary, we do not detach the normalization factor in Eq. \ref{eq:appendix_loss5}. This is intentional even if somewhat un-intuitive. We show in Figure \ref{fig:cub_car_ablations_detach} that detaching $|\R_{k/2 + \epsilon}(i)|$ in Eq. \ref{eq:appendix_loss5} increases R@1 on SOP when $\lambda$ is high, but decreases R@1 in all other scenarios.

\setlength\tabcolsep{4 pt}
\begin{table}
\caption{Similarity Regularizer Results. Here, we experiment with naively adding our similarity regularizer to a subset of the baseline methods. The overall loss is $\gamma \L_{\text{reg}} + (1-\gamma) (\text{baseline loss})$ where $\gamma=0.1$. We try two different values for $\tilde{s}$ and compare the difference in various performance metrics with the original baseline. The colored subscript denotes the increase or decrease in each metric when our similarity regularizer is added. Observe that in most cases, the regularizer improves the performance of the baseline or has negligible effect, with an appropriate $\tilde{s}$ value. }
\centering
{
\begin{tabular}{ l c c c c c c >{\columncolor[gray]{0.9}}c >{\columncolor[gray]{0.9}}c >{\columncolor[gray]{0.9}}c }
\toprule
& \multicolumn{9}{c}{Cars} \\
\midrule
& \multicolumn{3}{c}{mAP} & \multicolumn{3}{c}{mAP@R} & \multicolumn{3}{c}{R@1} \\
\cmidrule(lr){2-4} \cmidrule(lr){5-7} \cmidrule(lr){8-10}
Method & No $\L_{\text{reg}}$ & $\tilde{s} = 0.25$  & $\tilde{s} = 0.3$ & No $\L_{\text{reg}}$ & $\tilde{s} = 0.25$  & $\tilde{s} = 0.3$ & No $\L_{\text{reg}}$ & $\tilde{s} = 0.25$  & $\tilde{s} = 0.3$\\
\midrule
Contrastive  & 38.2 & 40.6  {\tiny \color{blue}{+2.4}} & 39.4 {\tiny \color{blue}{+1.2}} & 28.8 & 31.2  {\tiny \color{blue}{+2.4}} & 30.0  {\tiny \color{blue}{+1.2}} & 85.4 & 88.0  {\tiny \color{blue}{+2.6}} & 87.1  {\tiny \color{blue}{+1.7}} \\ 
Triplet  & 34.9 & 39.2  {\tiny \color{blue}{+4.3}} & 38.5 {\tiny \color{blue}{+3.6}} & 24.7 & 29.9  {\tiny \color{blue}{+5.2}} & 29.2  {\tiny \color{blue}{+4.5}} & 77.6 & 87.8  {\tiny \color{blue}{+10.2}} & 86.7  {\tiny \color{blue}{+9.1}} \\ 
NtXent  & 36.2 & 37.0  {\tiny \color{blue}{+0.8}} & 35.8 {\tiny \color{red}{-0.4}} & 26.3 & 27.0  {\tiny \color{blue}{+0.7}} & 26.0  {\tiny \color{red}{-0.3}} & 79.0 & 79.9  {\tiny \color{blue}{+0.9}} & 78.7  {\tiny \color{red}{-0.3}} \\ 
MS  & 43.1 & 42.7  {\tiny \color{red}{-0.4}} & \textbf{43.6} {\tiny \color{blue}{+0.5}} & \textbf{33.7} & 33.3  {\tiny \color{red}{-0.4}} & 34.1  {\tiny \color{blue}{+0.4}} & 88.7 & 88.8  {\tiny \color{blue}{+0.1}} & \textbf{89.3}  {\tiny \color{blue}{+0.6}} \\ 
Smooth-AP  & 37.5 & 38.9  {\tiny \color{blue}{+1.4}} & 38.1 {\tiny \color{blue}{+0.6}} & 27.5 & 28.8  {\tiny \color{blue}{+1.3}} & 28.2  {\tiny \color{blue}{+0.7}} & 81.1 & 82.1  {\tiny \color{blue}{+1.0}} & 82.4  {\tiny \color{blue}{+1.3}} \\ 
ROADMAP  & 38.3 & 39.2  {\tiny \color{blue}{+0.9}} & 39.6 {\tiny \color{blue}{+1.3}} & 28.7 & 29.6  {\tiny \color{blue}{+0.9}} & 30.0  {\tiny \color{blue}{+1.3}} & 84.5 & 85.7  {\tiny \color{blue}{+1.2}} & 85.7  {\tiny \color{blue}{+1.2}} \\ 
Ours & -- & 42.4 & -- & -- & 33.0 & -- & -- & \textbf{89.3} & --\\
\bottomrule
\end{tabular}

\begin{tabular}{ l c c c c c c >{\columncolor[gray]{0.9}}c >{\columncolor[gray]{0.9}}c >{\columncolor[gray]{0.9}}c }
\toprule
& \multicolumn{9}{c}{SOP} \\
\midrule
& \multicolumn{3}{c}{mAP} & \multicolumn{3}{c}{mAP@R} & \multicolumn{3}{c}{R@1} \\
\cmidrule(lr){2-4} \cmidrule(lr){5-7} \cmidrule(lr){8-10}
Method & No $\L_{\text{reg}}$ & $\tilde{s} = 0.25$  & $\tilde{s} = 0.3$ & No $\L_{\text{reg}}$ & $\tilde{s} = 0.25$  & $\tilde{s} = 0.3$ & No $\L_{\text{reg}}$ & $\tilde{s} = 0.25$  & $\tilde{s} = 0.3$\\
\midrule
Contrastive  & 62.9 & 62.8  {\tiny \color{red}{-0.1}} & 62.7 {\tiny \color{red}{-0.2}} & 57.0 & 56.9  {\tiny \color{red}{-0.1}} & 56.8  {\tiny \color{red}{-0.2}} & 82.4 & 82.3  {\tiny \color{red}{-0.1}} & 82.3  {\tiny \color{red}{-0.1}} \\ 
Triplet  & 63.1 & 64.4  {\tiny \color{blue}{+1.3}} & 64.6 {\tiny \color{blue}{+1.5}} & 56.4 & 57.8  {\tiny \color{blue}{+1.4}} & 58.0  {\tiny \color{blue}{+1.6}} & 82.0 & 83.1  {\tiny \color{blue}{+1.1}} & \textbf{83.3}  {\tiny \color{blue}{+1.3}} \\ 
NtXent  & 60.1 & 60.3  {\tiny \color{blue}{+0.2}} & 60.0 {\tiny \color{red}{-0.1}} & 53.4 & 53.7  {\tiny \color{blue}{+0.3}} & 53.4  {\tiny \color{blue}{+0.0}} & 79.7 & 80.0  {\tiny \color{blue}{+0.3}} & 79.7  {\tiny \color{blue}{+0.0}} \\ 
MS  & 62.5 & 62.4  {\tiny \color{red}{-0.1}} & 62.6 {\tiny \color{blue}{+0.1}} & 56.2 & 56.1  {\tiny \color{red}{-0.1}} & 56.3  {\tiny \color{blue}{+0.1}} & 81.4 & 81.4  {\tiny \color{blue}{+0.0}} & 81.8  {\tiny \color{blue}{+0.4}} \\ 
Smooth-AP  & 63.1 & 63.2  {\tiny \color{blue}{+0.1}} & 63.5 {\tiny \color{blue}{+0.4}} & 56.4 & 56.5  {\tiny \color{blue}{+0.1}} & 56.8  {\tiny \color{blue}{+0.4}} & 82.0 & 82.1  {\tiny \color{blue}{+0.1}} & 82.3  {\tiny \color{blue}{+0.3}} \\ 
ROADMAP  & 64.4 & 64.4  {\tiny \color{blue}{+0.0}} & 64.4 {\tiny \color{blue}{+0.0}} & 58.2 & 58.2  {\tiny \color{blue}{+0.0}} & 58.2  {\tiny \color{blue}{+0.0}} & 83.1 & 83.1  {\tiny \color{blue}{+0.0}} & 83.1  {\tiny \color{blue}{+0.0}} \\ 
Ours  & -- & \textbf{65.0} & -- & -- & \textbf{58.6} & -- & -- & \textbf{83.3} & -- \\
\bottomrule
\end{tabular}

\begin{tabular}{ l c c c c c c >{\columncolor[gray]{0.9}}c >{\columncolor[gray]{0.9}}c >{\columncolor[gray]{0.9}}c }
\toprule
& \multicolumn{9}{c}{iNaturalist} \\
\midrule
& \multicolumn{3}{c}{mAP} & \multicolumn{3}{c}{mAP@R} & \multicolumn{3}{c}{R@1} \\
\cmidrule(lr){2-4} \cmidrule(lr){5-7} \cmidrule(lr){8-10}
Method & No $\L_{\text{reg}}$ & $\tilde{s} = 0.25$  & $\tilde{s} = 0.3$ & No $\L_{\text{reg}}$ & $\tilde{s} = 0.25$  & $\tilde{s} = 0.3$ & No $\L_{\text{reg}}$ & $\tilde{s} = 0.25$  & $\tilde{s} = 0.3$\\
\midrule
Contrastive  & 16.0 & 15.1  {\tiny \color{red}{-0.9}} & 15.5 {\tiny \color{red}{-0.5}} & 11.6 & 11.2  {\tiny \color{red}{-0.4}} & 11.5  {\tiny \color{red}{-0.1}} & 43.5 & 43.4  {\tiny \color{red}{-0.1}} & 43.7  {\tiny \color{blue}{+0.2}} \\ 
Triplet  & 12.1 & 13.3  {\tiny \color{blue}{+1.2}} & 13.6 {\tiny \color{blue}{+1.5}} & 7.9 & 9.8  {\tiny \color{blue}{+1.9}} & 10.0  {\tiny \color{blue}{+2.1}} & 35.4 & 41.4  {\tiny \color{blue}{+6.0}} & 41.7  {\tiny \color{blue}{+6.3}} \\ 
NtXent  & 15.9 & 15.9  {\tiny \color{blue}{+0.0}} & 15.9 {\tiny \color{blue}{+0.0}} & 10.6 & 10.7  {\tiny \color{blue}{+0.1}} & 10.6  {\tiny \color{blue}{+0.0}} & 40.8 & 40.7  {\tiny \color{red}{-0.1}} & 40.7  {\tiny \color{red}{-0.1}} \\ 
MS  & 16.6 & 16.6  {\tiny \color{blue}{+0.0}} & 16.6 {\tiny \color{blue}{+0.0}} & 11.9 & 11.9  {\tiny \color{blue}{+0.0}} & 11.9  {\tiny \color{blue}{+0.0}} & 44.9 & 44.9  {\tiny \color{blue}{+0.0}} & 45.0  {\tiny \color{blue}{+0.1}} \\ 
Smooth-AP  & 16.5 & 16.6  {\tiny \color{blue}{+0.1}} & 16.7 {\tiny \color{blue}{+0.2}} & 11.3 & 11.4  {\tiny \color{blue}{+0.1}} & 11.4  {\tiny \color{blue}{+0.1}} & 42.7 & 43.1  {\tiny \color{blue}{+0.4}} & 43.2  {\tiny \color{blue}{+0.5}} \\ 
ROADMAP  & 17.8 & \textbf{18.1}  {\tiny \color{blue}{+0.3}} & 18.0 {\tiny \color{blue}{+0.2}} & 12.7 & \textbf{13.1}  {\tiny \color{blue}{+0.4}} & 13.0  {\tiny \color{blue}{+0.3}} & 45.9 & \textbf{47.1}  {\tiny \color{blue}{+1.2}} & 46.8  {\tiny \color{blue}{+0.9}} \\ 
Ours  & -- & 17.1 & -- & --& 12.4 & -- & -- & 46.2 & -- \\
\bottomrule
\end{tabular}
}
\label{tab:regsim-results}
\end{table}
\setlength\tabcolsep{6 pt}

\section{Additional Experimental Results}

\subsection{Experimental Results on Similarity Regularizer } In the main paper, we presented the similarity regularizer $\L_{\text{reg}}$ as complementary to the contextual similarity loss $\L_{\text{context}}$. However, there is no theoretical reason for limiting the regularizer to our contextual similarity framework. We show in Table \ref{tab:regsim-results} that the similarity regularizer offers an improvement in retrieval performance when combined with some baselines. In particular, the regularizer consistently improves R@1 performance of Triplet, Multi-Similarity, Smooth-AP, and ROADMAP losses across Cars, SOP and iNaturalist benchmarks. This is a promising result.

\subsection{In-Shop Results } We offer R@$k$ results for standard $k$ values on the In-Shop dataset in Table \ref{tab:inshop-results}. In-Shop (\cite{liu2016deepfashion}) is a popular image retrieval benchmark. We use the standard train-test split: we use 25,882 images from 3,997 classes for training; for testing, we use a query set with 14,218 images and a gallery set (aka. reference set) with 12,612 images, both from 3,985 classes. We use the same augmentation and batch size as SOP. An epoch is defined as 600 batches.

\subsection{Average Precision Results for Table \ref{tab:main} }  We offer Average Precision (AP) results in Table \ref{tab:AP-results}. We present results for two flavors of AP (\cite{mcfee2010metric}): mAP and mAP@R. mAP is the precision at $k$, averaged across correctly retrieved samples, averaged across the query set. mAP@R (\cite{musgrave2020metric}) only averages across retrievals up to the number of positive pairs in the gallery set. Hence, mAP@R is always smaller than mAP. The two versions of AP are equivalent bases for comparison. In the following equations, $\X^+_i$ denotes the set of samples in the gallery set with the same label as query $i$ and $\X^-_i$ denotes the set of samples in the gallery set with a different label than query $i$. $\text{Prec@$k$}$ denotes the precision at $k$, which is the percentage of samples ranked less than or equal to $k$ with the same label as the query.
\[
\text{mAP}_i = \frac{1}{|\X^+_i|} \sum_{k=1}^{|\X^+_i| + |\X^-_i|} \text{Prec@$k$} \: \ind [k \in \X^+_i]
   \]
\[
\text{mAP@R}_i = \frac{1}{|\X^+_i|} \sum_{k=1}^{|\X^+_i|} \text{Prec@$k$} \: \ind [k \in \X^+_i]
   \]

\begin{figure}
\centering
\begin{subfigure}{.32\textwidth}
  \centering
  \includegraphics[width=1.\linewidth]{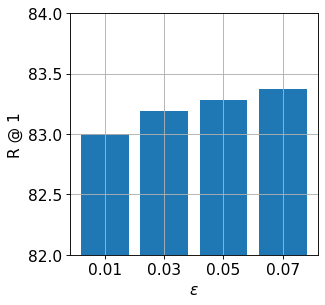}
\end{subfigure}
\begin{subfigure}{.32\textwidth}
  \centering
  \includegraphics[width=1.\linewidth]{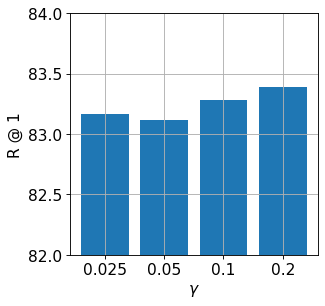}
\end{subfigure}
\begin{subfigure}{.32\textwidth}
  \centering
  \includegraphics[width=1.\linewidth]{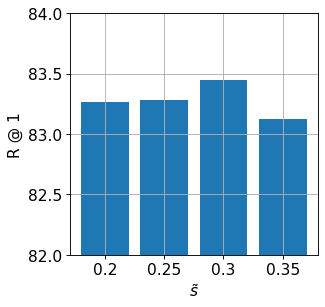}
\end{subfigure}
\caption{Hyperparameter tuning on SOP. We experiment with various values for $\epsilon$,  $\gamma$ and $\tilde{s}$. The R@1 values are similar for the different hyperparameter choices. }
\label{fig:hypers}
\end{figure}

\begin{figure}
\centering
\includegraphics[width=0.75\linewidth]{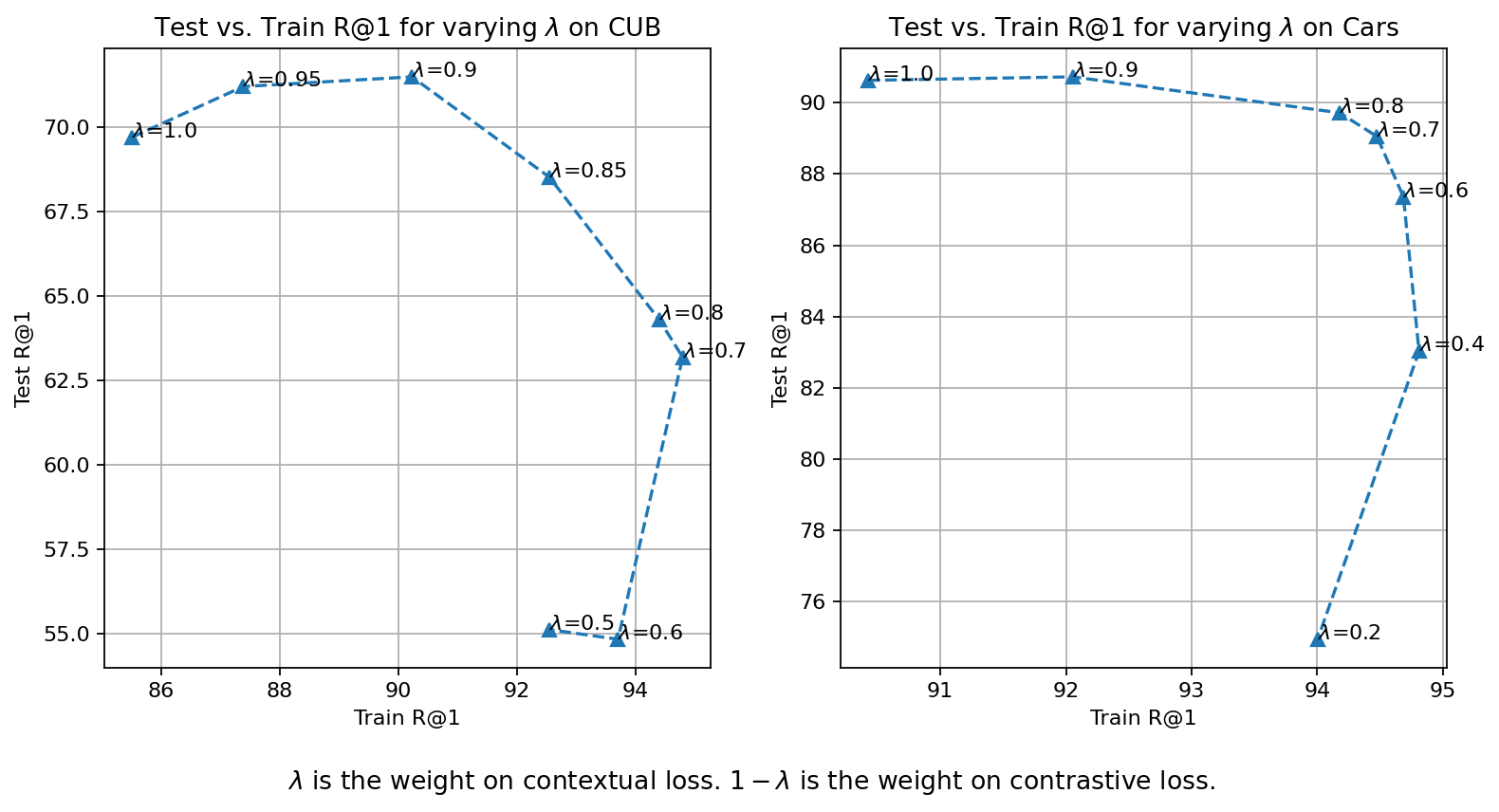}
\caption{Test R@1 vs. train R@1 accuracy for varying $\lambda$ ($\gamma=0$). This figure shows that adding the standard contrastive loss (lowering $\lambda$) increases the R@1 on training data. This also increases the test R@1 up to a point (approximately $\lambda=0.9$), before overfitting.}
\label{fig:not_overfitting}
\end{figure}

\begin{figure}
\centering
\begin{subfigure}{.5\textwidth}
  \centering
  \includegraphics[width=1.\linewidth]{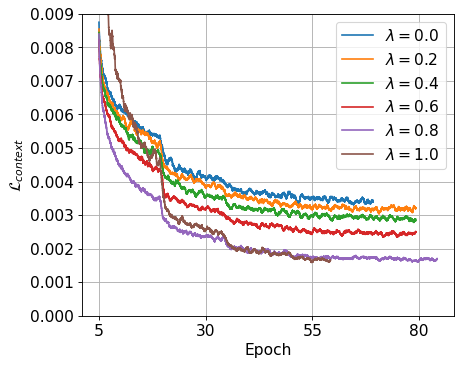}
\end{subfigure}
\begin{subfigure}{.46\textwidth}
  \centering
  \includegraphics[width=1.\linewidth]{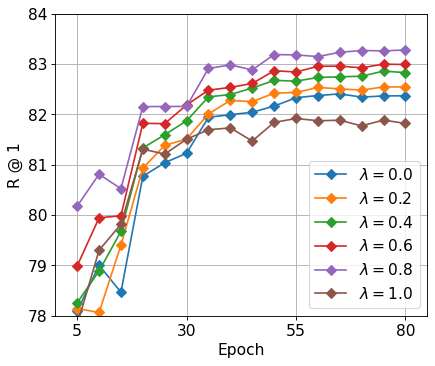}
\end{subfigure}
\caption{Plot of contextual similarity loss $\L_{\text{context}}$ with varying $\lambda$. Each color is a different $\lambda$. Results are on the SOP dataset. Observe that the contrastive loss implicitly minimizes the contextual loss (blue line). Also observe that in general, a lower contextual loss corresponds to a higher test R@1. This shows that our loss function is a reasonable objective. Note that the plot on the left plots $\L_{\text{context}}$, not the complete training loss. $\gamma=0$.}
\label{fig:loss_plot}
\end{figure}

\begin{figure}
\centering
\begin{subfigure}{.39\textwidth}
  \centering
  \includegraphics[width=1.\linewidth]{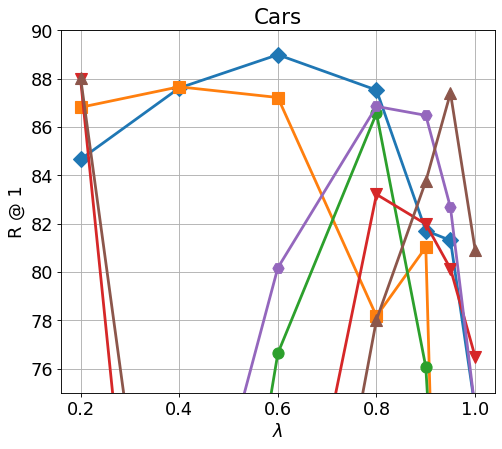}
\end{subfigure}
\begin{subfigure}{.59\textwidth}
  \centering
  \includegraphics[width=1.\linewidth]{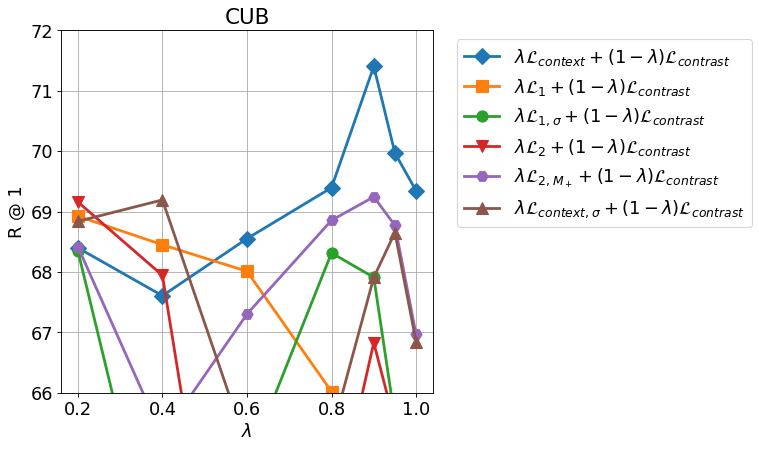}
\end{subfigure}
\caption{Ablation results on CUB and Cars. The blue line represents test R@1 of our framework for varying $\lambda$ with $\gamma=0$. The other colors represent various modifications or simplifications of the loss function (reference Table \ref{tab:ablation-2} and Section F for full description). We perform each ablation experiment with varying $\lambda$ since the optimal $\lambda$ is not constant across all experiments. Clearly, the modified versions of $\L_{\text{context}}$ are sub-optimal. }
\label{fig:cub_car_ablations}
\end{figure}

\begin{figure}
\centering
\begin{subfigure}{.31\textwidth}
  \centering
  \includegraphics[width=1.\linewidth]{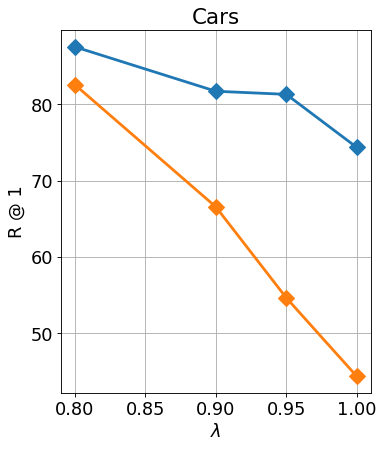}
\end{subfigure}
\begin{subfigure}{.32\textwidth}
  \centering
  \includegraphics[width=1.\linewidth]{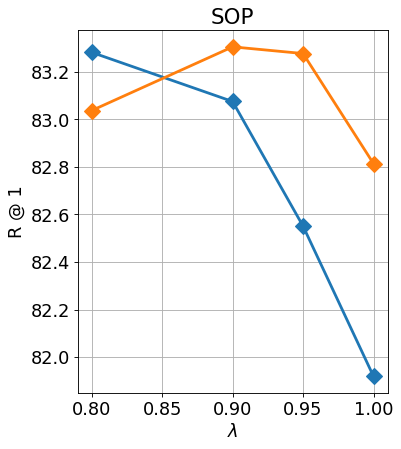}
\end{subfigure}
\begin{subfigure}{.31\textwidth}
  \centering
  \includegraphics[width=1.\linewidth]{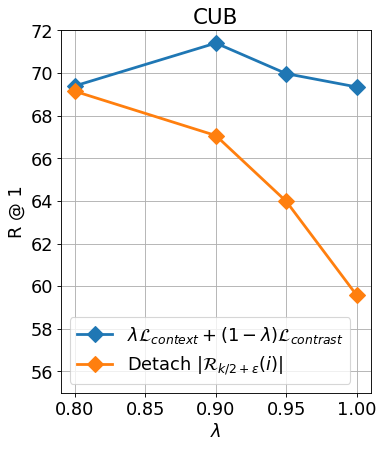}
\end{subfigure}
\caption{Investigating the effect of detaching $|\R_{k/2 + \epsilon}(i)|$ in Eq. \ref{eq:appendix_loss5}. The blue line shows R@1 of our framework with varying $\lambda$. $\gamma=0$. The orange line shows R@1 after detaching $|\R_{k/2 + \epsilon}(i)|$. Overall, detaching this normalization factor is undesirable.}
\label{fig:cub_car_ablations_detach}
\end{figure}

\setlength\tabcolsep{3 pt}
\begin{table}
\caption{mAP and mAP@R Results. We use ResNet-50 with an embedding size of 512 for all experiments. We re-run all baselines in this Table using the implementation by \cite{musgrave2020pytorch}. Standard deviations are based on three trials with the same train-test split. } 
\centering
{  
\small
\begin{tabular}{ l c c c c c c c c  }
 \toprule
 & \multicolumn{2}{c}{CUB} & \multicolumn{2}{c}{Cars} & \multicolumn{2}{c}{SOP}& \multicolumn{2}{c}{mini-iNaturalist}\\
 \midrule
 Method & mAP & mAP@R & mAP & mAP@R & mAP & mAP@R & mAP & mAP@R  \\
  \midrule
Contrastive & 36.6 $\pm$ 0.4 & 26.6 $\pm$ 0.5 & 38.2 $\pm$ 0.4 & 28.8 $\pm$ 0.5 & 62.9 $\pm$ 0.1 & 57.0 $\pm$ 0.1 & 16.0 $\pm$ 0.1 & 11.6 $\pm$ 0.0 \\
 Triplet & 36.9 $\pm$ 0.4 & 26.5 $\pm$ 0.2 & 34.9 $\pm$ 0.8 & 24.7 $\pm$ 0.7 & 63.1 $\pm$ 0.0 & 56.4 $\pm$ 0.0 & 12.1 $\pm$ 0.0 & 7.9 $\pm$ 0.0 \\
 NtXent & 36.3 $\pm$ 0.3 & 26.0 $\pm$ 0.3 & 36.2 $\pm$ 0.7 & 26.3 $\pm$ 0.7 & 60.1 $\pm$ 0.1 & 53.4 $\pm$ 0.1 & 15.9 $\pm$ 0.0 & 10.6 $\pm$ 0.0 \\
 MS & 38.0 $\pm$ 0.1 & 27.7 $\pm$ 0.1 & \textbf{43.1} $\pm$ 0.3 & \textbf{33.7} $\pm$ 0.3 & 62.5 $\pm$ 0.0 & 56.2 $\pm$ 0.1 & 16.6 $\pm$ 0.0 & 11.9 $\pm$ 0.0 \\
 Fast-AP & 34.4 $\pm$ 0.8 & 24.4 $\pm$ 0.7 & 32.7 $\pm$ 0.0 & 23.5 $\pm$ 0.0 & 60.7 $\pm$ 0.1 & 54.2 $\pm$ 0.1 & 13.9 $\pm$ 0.1 & 9.1 $\pm$ 0.1 \\
 Smooth-AP & 36.8 $\pm$ 0.4 & 26.4 $\pm$ 0.4 & 37.5 $\pm$ 0.3 & 27.5 $\pm$ 0.2 & 63.1 $\pm$ 0.2 & 56.4 $\pm$ 0.2 & 16.5 $\pm$ 0.0 & 11.3 $\pm$ 0.0 \\
 ROADMAP & 37.7 $\pm$ 0.1 & 27.5 $\pm$ 0.2 & 38.3 $\pm$ 0.4 & 28.7 $\pm$ 0.4 & 64.4 $\pm$ 0.1 & 58.2 $\pm$ 0.1 & \textbf{17.8} $\pm$ 0.0 & \textbf{12.7} $\pm$ 0.0 \\
 Ours & \textbf{38.3} $\pm$ 0.2 & \textbf{28.0} $\pm$ 0.2 & 42.4 $\pm$ 0.4 & 33.0 $\pm$ 0.3 & \textbf{65.0} $\pm$ 0.1 & \textbf{58.6} $\pm$ 0.1 & 17.1 $\pm$ 0.0 & 12.4 $\pm$ 0.0 \\
 \bottomrule
\end{tabular}
}
\label{tab:AP-results}
\end{table}
\setlength\tabcolsep{6 pt}

\begin{table}
\caption{In-Shop Results. We use ResNet-50 with an embedding size of 512 for all experiments. $\dagger$ indicates results reported by the original authors; we re-run all other baselines using the implementation by \cite{musgrave2020pytorch}. Our R@1 performance is better than all baselines. We are at least comparable to baselines on other R@$k$ metrics.} 
\centering
{
\begin{tabular}{ l >{\columncolor[gray]{0.9}}c c c c c c }
 \toprule
 & \multicolumn{6}{c}{In-Shop} \\
 \midrule
Method & R@1 & R@10 & R@20 & R@30 & R@40 & R@50 \\
 \midrule
 Contrastive & 90.1 & 97.4 & 98.3 & 98.6 & 98.8 & 98.9 \\
 Triplet & 90.2 & 98.0 & 98.7 & \textbf{99.0} & \textbf{99.2} & \textbf{99.3} \\
 NtXent & 89.3 & 97.6 & 98.3 & 98.7 & 98.9 & 99.0 \\
 MS & 86.9 & 96.1 & 97.3 & 97.9 & 98.1 & 98.3 \\
 N-Softmax$\dagger$ & 88.6 & 97.5 & 98.4 & 98.8 & -- & --  \\
 Proxy NCA ++$\dagger$ & 90.4 & \textbf{98.1} & \textbf{98.8} & \textbf{99.0} & \textbf{99.2} & -- \\
 Fast-AP & 89.7 & 97.5 & 98.3 & 98.6 & 98.8 & 98.9 \\
 Smooth-AP & 90.2 & 97.9 & 98.7 & \textbf{99.0} & \textbf{99.2} & 99.2 \\
 ROADMAP & 90.4 & 97.6 & 98.3 & 98.6 & 98.8 & 99.0 \\
 Ours & \textbf{90.7} & 97.8 & 98.5 & 98.9 & 99.1 & 99.2 \\
 \bottomrule
\end{tabular}
}
\label{tab:inshop-results}
\end{table}

\subsection{Comparison on Hierarchical Retrieval Metrics}
Over-reliance on binary supervision is a long-standing problem in metric learning which motivates our contextual loss. Another interesting line of work uses hierarchical labels during training to mitigate this over-reliance (\citet{sun2021dynamic}, \citet{zheng2022dynamic}, and \citet{ramzi2022hierarchical}). These works fall under the umbrella of ``dynamic metric learning'' and ``hierarchical metric learning''. One potential drawback of these works is their requirement for hierarchical labels, which may not always be available. Broadly speaking, hierarchical metric learning losses enforce a larger penalty on mistakes in discriminating between labels that are farther apart in the hierarchy during training. The resulting embedding space is more consistent in the sense that most retrieval mistakes are between labels close together in the hierarchy. For example, if the query is an image of a bird, any incorrectly retrieved images are likely to be birds from a closely related species, rather than images of other animals.

The contextual loss as presented in this paper does not use hierarchical labels, since our focus is advancing the state-of-the-art in fine-grained retrieval, irrespective of coarse-grained retrieval. Keeping this in mind, we offer results on hierarchical retrieval metrics in Table \ref{tab:hierarchicalmetrics}. From Table \ref{tab:hierarchicalmetrics}, we conclude that our contextual loss always achieves the highest fine-grained performance, but under-performs baselines at higher levels in the hierarchy. This trade-off in fine vs. coarse retrieval is consistent with results from HAPPIER \citep{ramzi2022hierarchical}. We hope that these results will be helpful for future work on adapting the contextual loss to the hierarchical retrieval setting.

\setlength\tabcolsep{2.5 pt}
\begin{table}
\caption{ Hierarchical Retrieval Results. We use ResNet-50 with an embedding size of 512 for all experiments. We re-run all baselines in this table using the implementation by \cite{musgrave2020pytorch}. Image size is 224x224, and the experimental settings follow the settings presented in Section 5 of the main paper. The results in this table are from one random trial. ``fine R@1'' indicates the R@1 at the fine-grained label level (which is the same R@1 as the rest of the paper). On CUB, there are two higher levels in the label hierarchy: the family and order that the bird species belongs to, indicated as ``mid'' and ``coarse'', resp. On Cars, the coarse label is the make of the vehicle. On SOP, the coarse label is the category of the product. We follow the hierarchical labels used in \citep{chang2021your} for CUB and Cars; we use the coarse labels given by the original SOP dataset. }
\centering
{
\begin{tabular}{l >{\columncolor[gray]{0.9}}c c c c >{\columncolor[gray]{0.9}}c c c >{\columncolor[gray]{0.9}}c c c}
\toprule
& \multicolumn{4}{c}{CUB} & \multicolumn{3}{c}{Cars} & \multicolumn{3}{c}{SOP} \\
\cmidrule(lr){2-5} \cmidrule(lr){6-8} \cmidrule(lr){9-11}
Method & fine R@1 & fine AP & mid AP & coarse AP & fine R@1 & fine AP & coarse AP & fine R@1 & fine AP & coarse AP \\ 
\midrule
Contrastive & 66.2 & 35.9 & 52.0 & 85.7 & 81.1 & 33.4 & 39.8 & 80.7 & 60.1 & 13.0 \\
Roadmap & 66.2 & 36.0 & 53.8 & 88.1 & 83.5 & 37.3 & 41.9 & 82.1 & 62.6 & 13.9 \\
Triplet & 65.0 & 34.5 & 53.8 & \textbf{89.8} & 89.3 & 43.3 & \textbf{42.7} & 81.8 & 62.5 & 15.0 \\
MS+miner & 69.4 & 37.4 & 53.5 & 88.5 & 90.7 & 42.7 & 39.4 & 82.1 & 63.2 & 13.5 \\
Proxy Anchor & 68.0 & 36.6 & 53.9 & 89.1 & 88.8 & 39.0 & 38.6 & 79.8 & 59.4 & 16.2 \\
Proxy NCA & 65.7 & 34.9 & \textbf{55.5} & 88.0 & 88.0 & 37.6 & 38.5 & 78.9 & 58.6 & \textbf{16.8} \\
Contextual (ours) & \textbf{72.7} & \textbf{40.3} & 54.3 & 86.4 & \textbf{91.0} & \textbf{43.4} & 39.1 & \textbf{82.7} & \textbf{63.7} & 13.5 \\ 
\bottomrule
\end{tabular}
}
\label{tab:hierarchicalmetrics}
\end{table}
\setlength\tabcolsep{6 pt}

\subsection{Contextual Similarity Re-ranking at Test Time}
Using contextual similarity for re-ranking at test time is a common trick used to improve recall accuracy. However, this is not a common evaluation setting, so we omitted it from the main paper. We offer R@1 results with contextual re-ranking in Table \ref{tab:contextual-reranking}. The contextual re-ranking procedure is as follows. After training, we calculate the cosine similarity between all pairs of samples in the test set. Denote this as $s_{ij}$, for each pair of test samples $i$ and $j$. We then calculate a \emph{simplified version} of the contextual similarity $\tilde{w}_{ij}$ between each pair of test samples $i$ and $j$ using Equations \ref{eq:appendix_contextsim1} and \ref{eq:appendix_contextsim2} in Appendix E. Note that $s_{ij} \in [-1, 1]$ while $\tilde{w}_{ij} \in [0,1]$, so we need to rectify $s_{ij}$ to occupy the same interval as  $\tilde{w}_{ij}$. To this end, define $\tilde{s}_{ij} = \exp(2s_{ij} - 2)$. For retrieval, we use a weighted sum of the rectified cosine similarity and the contextual similarity: $\beta \tilde{w}_{ij} + (1-\beta) \tilde{s}_{ij} $.

Table \ref{tab:contextual-reranking} shows that all methods benefit from contextual re-ranking, and our contextual loss remains the best method across the three benchmarks. Embeddings optimized by the contextual loss do not benefit more from contextual re-ranking than baselines (i.e. the relative increase in R@1 with contextual re-ranking is about the same across all methods). There is no theory to suggest that optimizing contextual similarity at train time results in better contextual similarity re-ranking at test time, and we make no claims regarding this.

\setlength\tabcolsep{6 pt}
\begin{table}
\caption{Contextual re-ranking R@1 results. $\beta=0.1$ is weight on contextual similarity. We experiment with different values for $k$. We use ResNet-50 with an embedding size of 512 for all experiments. We re-run all baselines in this table using the implementation by \cite{musgrave2020pytorch}. Image size is 224x224, and the experimental settings follow the settings presented in Section 5 of the main paper. The results in this table are from one random trial. }
\centering
{
\begin{tabular}{l c c c c c c c c c}
\toprule
 & \multicolumn{3}{c}{CUB} & \multicolumn{3}{c}{Cars} & \multicolumn{2}{c}{SOP} \\
\cmidrule(lr){2-4} \cmidrule(lr){5-7} \cmidrule(lr){8-9}
Method & no re-rank & $k=16$ & $k=32$ & no re-rank & $k=16$ & $k=32$ & no re-rank & $k=4$  \\
\midrule
Contrastive & 66.2 & 66.6 & 66.8 & 81.1 & 81.2 & 80.8 & 80.7 & 81.0 \\
Roadmap & 66.2 & 67.4 & 66.9 & 83.5 & 83.5 & 83.3 & 82.1 & 82.3 \\
Triplet & 65.0 & 66.6 & 65.8 & 89.3 & 90.0 & 89.9 & 81.8 & 82.2 \\
MS+miner & 69.4 & 71.1 & 70.7 & 90.7 & 91.2 & 91.0 & 82.1 & 82.4 \\
Proxy Anchor & 68.0 & 69.1 & 69.0 & 88.8 & 89.3 & 89.1 & 79.8 & 80.1 \\
Proxy NCA & 65.7 & 67.2 & 66.9 & 88.0 & 88.5 & 88.5 & 78.9 & 79.1 \\
Contextual (ours) & \textbf{72.7} & \textbf{74.3} & \textbf{73.9} & \textbf{91.0} & \textbf{91.4} & \textbf{91.1} & \textbf{82.7} & \textbf{82.9} \\
\bottomrule
\end{tabular}
}
\label{tab:contextual-reranking}
\end{table}

\subsection{Additional Comparisons to AP Surrogates}
\begin{figure}
\centering
\includegraphics[width=1.\linewidth]{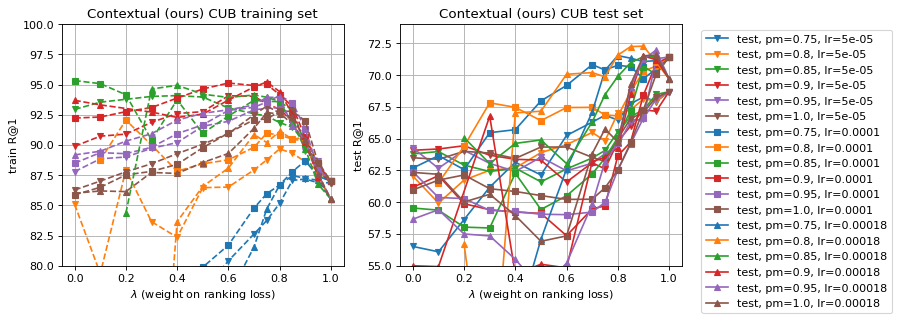}
\caption{CUB R@1 results for our contextual loss function, with varying $\lambda$, learning rate, and positive margin on contrastive loss. We plot both train and test R@1 results. ``lr'' and ``pm'' in the legend stand for learning rate and positive margin $\delta_+$, respectively.  }
\label{fig:lambda_cub_hybrid}
\end{figure}

\begin{figure}
\centering
\includegraphics[width=1.\linewidth]{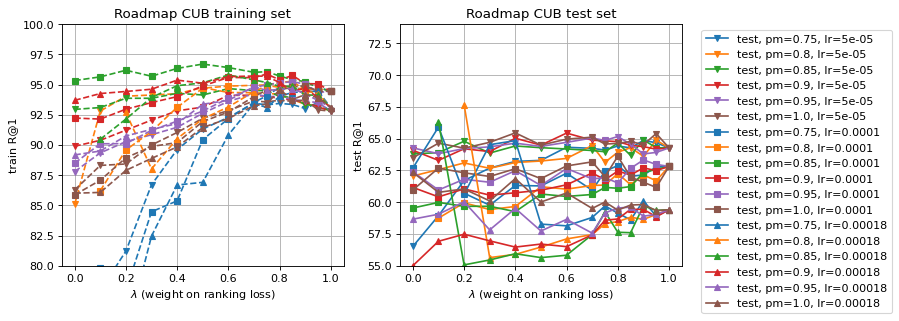}
\caption{CUB R@1 results for Roadmap, with varying $\lambda$, learning rate, and positive margin on contrastive loss. We plot both train and test R@1 results. ``lr'' and ``pm'' in the legend stand for learning rate and positive margin $\delta_+$, respectively.  }
\label{fig:lambda_cub_roadmap}
\end{figure}

\begin{figure}
\centering
\includegraphics[width=1.\linewidth]{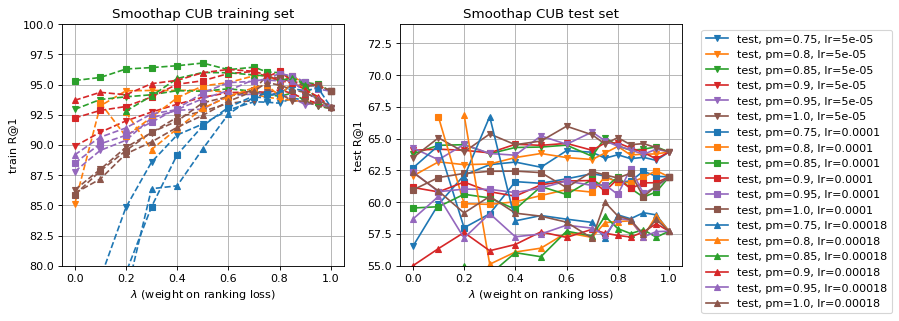}
\caption{CUB R@1 results for Smooth-AP, with varying $\lambda$, learning rate, and positive margin on contrastive loss. We plot both train and test R@1 results. ``lr'' and ``pm'' in the legend stand for learning rate and positive margin $\delta_+$, respectively.  }
\label{fig:lambda_cub_smoothap}
\end{figure}

\begin{figure}
\centering
\includegraphics[width=1.\linewidth]{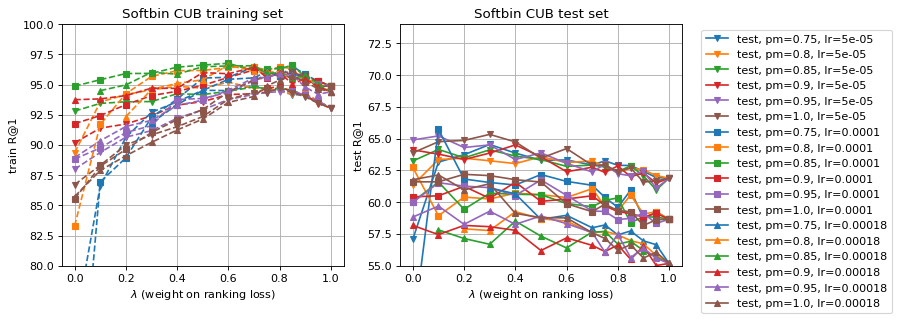}
\caption{CUB R@1 results for Softbin-AP, with varying $\lambda$, learning rate, and positive margin on contrastive loss. We plot both train and test R@1 results. ``lr'' and ``pm'' in the legend stand for learning rate and positive margin $\delta_+$, respectively.  }
\label{fig:lambda_cub_softbin}
\end{figure}

\begin{figure}
\centering
\includegraphics[width=1.\linewidth]{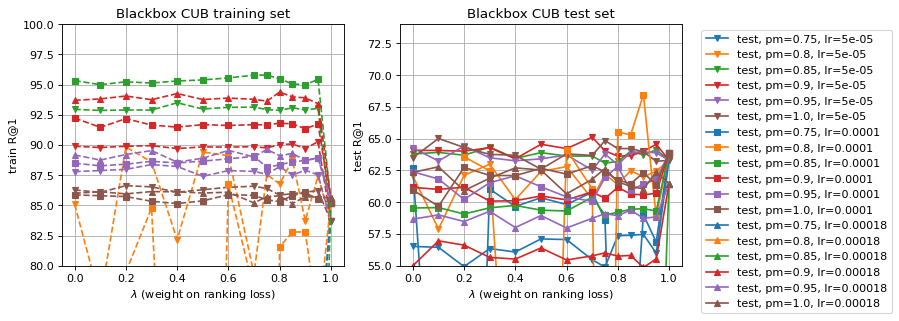}
\caption{CUB R@1 results for Blackbox-AP, with varying $\lambda$, learning rate, and positive margin on contrastive loss. We plot both train and test R@1 results. ``lr'' and ``pm'' in the legend stand for learning rate and positive margin $\delta_+$, respectively.  }
\label{fig:lambda_cub_blackbox}
\end{figure}

\begin{figure}
\centering
\includegraphics[width=1.\linewidth]{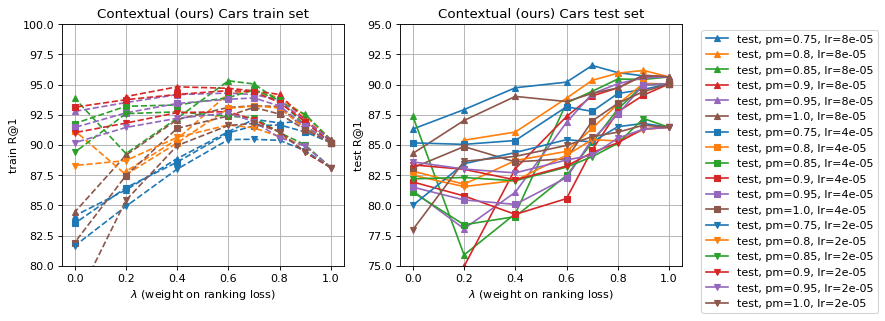}
\caption{Cars R@1 results for our contextual loss function, with varying $\lambda$, learning rate, and positive margin on contrastive loss. We plot both train and test R@1 results. ``lr'' and ``pm'' in the legend stand for learning rate and positive margin $\delta_+$, respectively.  }
\label{fig:lambda_cars_hybrid}
\end{figure}

\begin{figure}
\centering
\includegraphics[width=1.\linewidth]{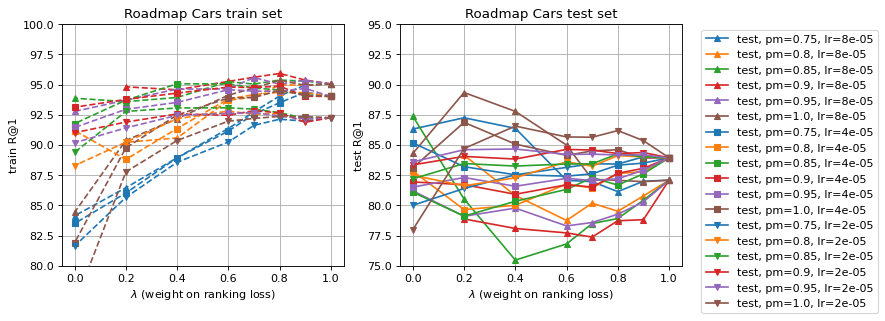}
\caption{Cars R@1 results for Roadmap, with varying $\lambda$, learning rate, and positive margin on contrastive loss. We plot both train and test R@1 results. ``lr'' and ``pm'' in the legend stand for learning rate and positive margin $\delta_+$, respectively.  }
\label{fig:lambda_cars_roadmap}
\end{figure}

\begin{figure}
\centering
\includegraphics[width=1.\linewidth]{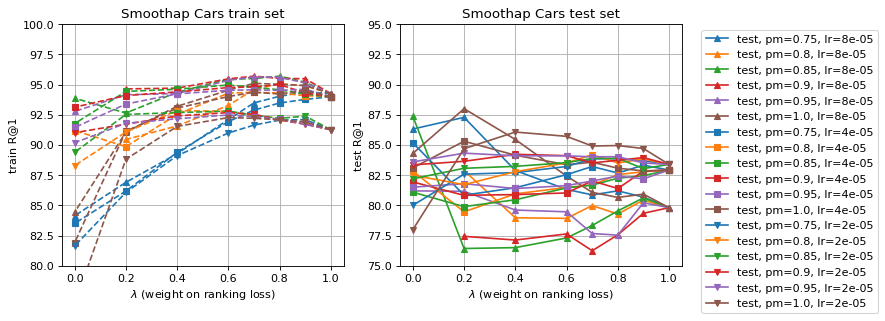}
\caption{Cars R@1 results for Smooth-AP, with varying $\lambda$, learning rate, and positive margin on contrastive loss. We plot both train and test R@1 results. ``lr'' and ``pm'' in the legend stand for learning rate and positive margin $\delta_+$, respectively.  }
\label{fig:lambda_cars_smoothap}
\end{figure}

\begin{figure}
\centering
\includegraphics[width=1.\linewidth]{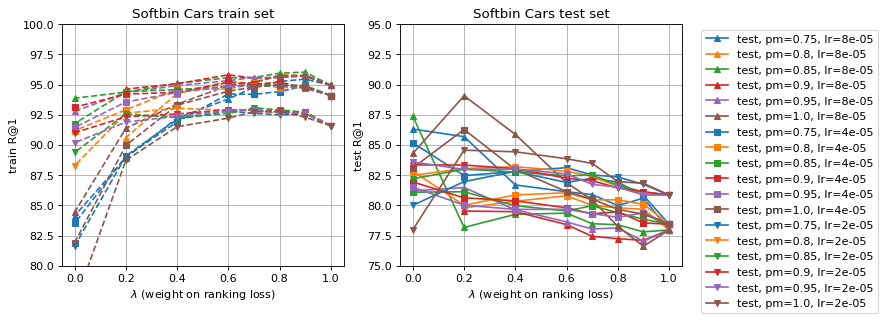}
\caption{Cars R@1 results for Softbin-AP, with varying $\lambda$, learning rate, and positive margin on contrastive loss. We plot both train and test R@1 results. ``lr'' and ``pm'' in the legend stand for learning rate and positive margin $\delta_+$, respectively.  }
\label{fig:lambda_cars_softbin}
\end{figure}

\begin{figure}
\centering
\includegraphics[width=1.\linewidth]{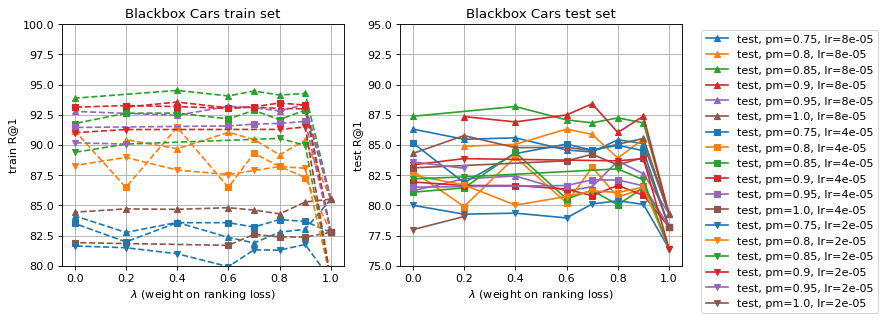}
\caption{Cars R@1 results for Blackbox-AP, with varying $\lambda$, learning rate, and positive margin on contrastive loss. We plot both train and test R@1 results. ``lr'' and ``pm'' in the legend stand for learning rate and positive margin $\delta_+$, respectively.  }
\label{fig:lambda_cars_blackbox}
\end{figure}

\begin{figure}
\centering
\includegraphics[width=1.\linewidth]{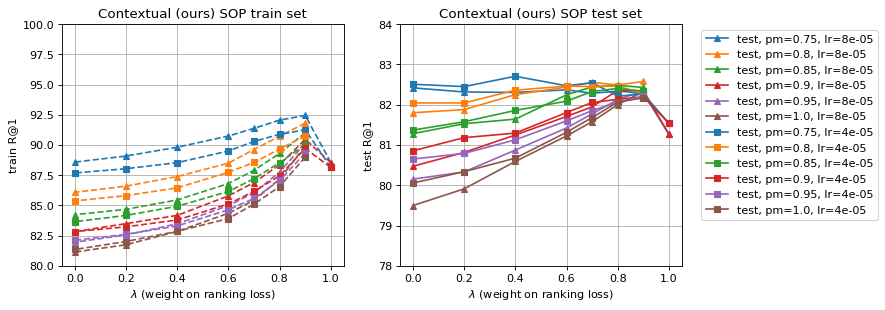}
\caption{SOP R@1 results for our contextual loss function, with varying $\lambda$, learning rate, and positive margin on contrastive loss. We plot both train and test R@1 results. ``lr'' and ``pm'' in the legend stand for learning rate and positive margin $\delta_+$, respectively.  }
\label{fig:lambda_sop_hybrid}
\end{figure}

\begin{figure}
\centering
\includegraphics[width=1.\linewidth]{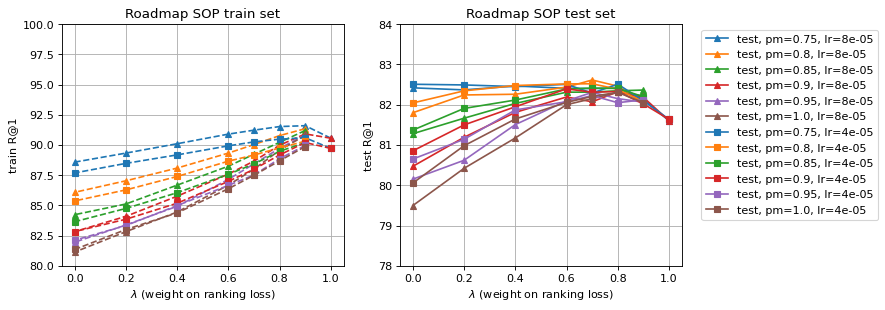}
\caption{SOP R@1 results for Roadmap, with varying $\lambda$, learning rate, and positive margin on contrastive loss. We plot both train and test R@1 results. ``lr'' and ``pm'' in the legend stand for learning rate and positive margin $\delta_+$, respectively.  }
\label{fig:lambda_sop_roadmap}
\end{figure}

Figures \ref{fig:lambda_cub_hybrid} through \ref{fig:lambda_sop_roadmap} present a comparison between our contextual loss function and various AP surrogates. The goal of these figures is to show that our contextual loss function is definitively better at ranking than AP surrogates. We optimize a convex combination of the ranking loss and standard contrastive loss. We present results for different choices of learning rates, $\lambda$ (weight on ranking loss), and positive margin on the contrastive loss, since the optimal hyperparameters vary between methods. We include results for all three benchmarks. The AP surrogates included in this comparison are: Roadmap \citep{ramzi2021robust}, Smooth-AP \cite{brown2020smooth}, Softbin-AP \citep{revaud2019learning}, and Blackbox-AP \citep{rolinek2020optimizing}.

\end{document}